\PassOptionsToPackage{table,dvipsnames}{xcolor}
\documentclass[11pt,a4paper,onecolumn]{tmlrmel}

\usepackage[numbers,sort&compress,square]{natbib}
\usepackage{amsmath,amsfonts,bm}

\def\eqref#1{equation~\ref{#1}}

\def\1{\bm{1}}

\DeclareMathAlphabet{\mathsfit}{\encodingdefault}{\sfdefault}{m}{sl}
\SetMathAlphabet{\mathsfit}{bold}{\encodingdefault}{\sfdefault}{bx}{n}

\usepackage{times}
\usepackage{amsmath,amssymb}
\usepackage{multirow}
\usepackage{wrapfig}
\usepackage{soul}
\usepackage{xspace}
\tcbuselibrary{most}
\newcommand{\cmark}{\ding{51}}
\newcommand{\xmark}{\ding{55}}
\newcommand{\best}[1]{\textbf{#1}}
\newcommand{\second}[1]{\underline{#1}}

\newcommand{\soap}{\textsc{Soap}\xspace}
\newcommand{\traj}{\tau}
\newcommand{\Sbase}{S}
\newcommand{\Stilde}{\widetilde{S}}
\newcommand{\proxy}{\mathcal{M}}
\newcommand{\ms}[2]{\ensuremath{#1_{\,\pm #2}}}

\definecolor{greyC}{RGB}{180,180,180}
\definecolor{greyL}{RGB}{235,235,235}
\definecolor{Gray}{gray}{0.9}

\definecolor{mydarkred}{rgb}{0.6,0,0}
\definecolor{myblue}{HTML}{268BD2}
\definecolor{mygreen}{HTML}{658354}
\definecolor{orangeinplot}{HTML}{e29c7a}
\definecolor{purpleinplot}{HTML}{7676a4}
\definecolor{greeninplot}{HTML}{288308}
\definecolor{mydarkblue}{rgb}{0,0.08,0.45}

\newtcolorbox{methodbox}{
  colback=black!2,
  colframe=black!55,
  boxrule=0.5pt,
  arc=0pt,
  left=5pt,
  right=5pt,
  top=4pt,
  bottom=4pt
}

\title{Can Failed Trajectories Reveal Where Agents Go Wrong? Agentic Failure Attribution Without Step-Level Labels}

\author[1,$\dagger$]{Do Nguyen-Thanh}
\author[1,$\dagger$]{Junlin Fang}
\author[2]{Bing Li}
\author[3]{Zhen Fang}
\author[1,*]{Sean Du}

\affil[1]{College of Computing and Data Science, Nanyang Technological University, Singapore}
\affil[2]{School of Electrical and Electronic Engineering, Nanyang Technological University, Singapore}
\affil[3]{Australian Artificial Intelligence Institute, University of Technology Sydney}
\affil[$\dagger$]{Equal contribution}
\affil[*]{Corresponding author}

\correspondingauthors{%
  \raggedright\mbox{\texttt{thanhdo001@e.ntu.edu.sg}}; 
  \mbox{\texttt{xuefeng.du@ntu.edu.sg}}
}

\coderepo{October, 2026}

\begin{abstract}

Failure attribution for LLM agents aims to identify  decisive steps responsible for task failure, which is critical for debugging and improving agentic systems. Existing methods rely on costly LLM prompting, step-level error annotations, or verified successful trajectories that lack failure signals for assistance. In this paper, we study failure attribution from \emph{step-unlabeled failure trajectories}, where each trajectory is known to have failed but its decisive step is unknown. We propose \soap (\textbf{S}pectral sc\textbf{O}ring with
 \textbf{A}ttention-guided \textbf{P}ropagation), an effective framework requiring neither step-level training labels nor a separate success-only reference set. \soap first constructs a spectral reference space from LLM representations and assigns each step a base error score according to its projection behavior. Importantly, later steps may exhibit stronger failure signals because they inherit an earlier mistake, causing independent scorers to favor downstream consequences over the decisive error. To address this \emph{downstream contamination}, \soap derives soft step dependencies from attention patterns and propagates downstream error evidence toward its likely upstream sources. Extensive experiments show that  \soap consistently improves attribution performance over competitive baselines.

\end{abstract}

\begin{document}
\maketitle

\section{Introduction}
\label{sec:intro}

When an LLM-based agent fails on a long-horizon task, identifying \emph{where the failure began} is essential for debugging, auditing, and improving the agent.
An agent trajectory may contain dozens or even hundreds of reasoning, communication, and tool-use steps, while only one early decision introduces the error that ultimately causes task failure.
Manually tracing such trajectories is prohibitively expensive and prone to inconsistency, especially as agents are increasingly deployed for software engineering, scientific discovery, and other complex tasks~\citep{yao2023react,wang2024surveyagents}.
This motivates the problem of \emph{failure attribution}: given a failed trajectory, identify the decisive step that introduced the failure~\citep{whoandwhen}.

Failure attribution is fundamentally challenging.
Existing methods typically rely either on prompting powerful LLM judges, which incurs substantial inference cost and can be unreliable over long trajectories~\citep{whoandwhen,echo,chief,falat,correct}, or on post-training attribution models using failure trajectories with step-level annotations~\citep{agentracer,graphtracer}.
Such annotations are costly and often ambiguous, since identifying the decisive step may require substantial domain expertise.
OAT~\citep{oat} alleviates this burden by learning the latent dynamics of successful trajectories and detecting deviations from the learned flow.
However, it has no access to failure trajectories during training, and therefore \textit{models only what successful behavior looks like}.
Moreover, its success-only reference set must be verified and may become stale as agents and task distributions evolve.
This motivates a complementary question:

\vspace{-0.5em}
\begin{center}
\textbf{Can failed trajectories themselves enable failure attribution
without step-level annotations?}
\vspace{-0.5em}
\end{center}

Failed trajectories provide a complementary source of supervision.
In many agentic systems, whether a task has failed can be determined from execution outcomes or task evaluators, while identifying the step responsible for that failure remains costly.
We therefore consider a collection of \emph{step-unlabeled failure trajectories}, i.e.,  every trajectory is known to have failed, but no step is annotated as decisive or non-decisive.
Each trajectory consequently contains an unlabeled mixture of ordinary steps and the decisive error.
Unlike OAT~\citep{oat}, such a failure-conditioned dataset directly exposes the behaviors that arise when agents fail.
Yet harnessing this information is non-trivial: without step-level membership labels, the learner must extract a meaningful failure signal without knowing which step introduced the failure.
The first challenge is therefore to \emph{identify decisive-error evidence from a step-level unlabeled mixture within failed trajectories}.

\begin{wrapfigure}{r}{0.48\linewidth}
\vspace{-0.9\baselineskip}
\centering
\includegraphics[width=\linewidth]{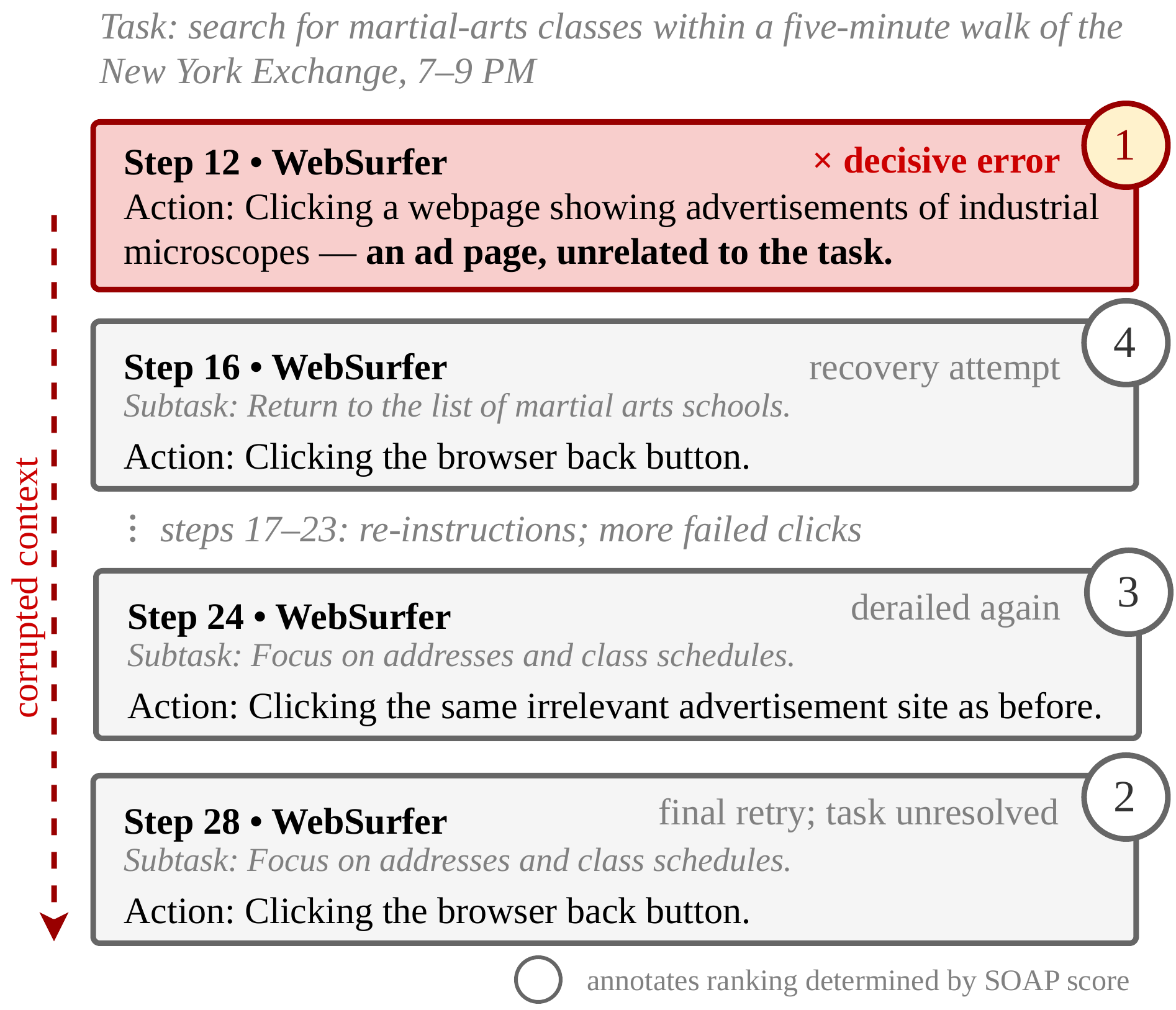}
\caption{\small \textbf{Downstream contamination} in a failed \textsc{Who\&When} trajectory~\citep{whoandwhen}. An early wrong click (step 12) lands the browser on an irrelevant page; later steps are spent on recovery attempts that derail again.}
\label{fig:contamination-example}
\vspace{-\baselineskip}
\end{wrapfigure}
A second challenge is that the step exhibiting the strongest failure signal need not be the step that introduced the failure. Agent trajectories are autoregressive because each step is generated from preceding observations and actions. Once an error occurs, later steps operate on a corrupted context and may produce failed tool calls, contradictory reasoning, or repeated recovery attempts. These downstream consequences can be more conspicuous than the original error. For example, a plausible but incorrect tool argument may trigger a sequence of visible error messages and unsuccessful retries, causing a conventional per-step scorer to select the final retry rather than the initial incorrect action. We refer to this phenomenon as \emph{downstream contamination} (Figure~\ref{fig:contamination-example}). Recent methods reconstruct dependencies through LLM prompting or specialized training, then perform graph search or backtracking~\citep{falat,chief,graphtracer}, which introduce substantial inference or training overhead. This motivates a lightweight alternative that extracts soft dependencies from the model to separate root causes from downstream effects.

In this paper, we propose \soap (\textbf{S}pectral sc\textbf{O}ring with
 \textbf{A}ttention-guided \textbf{P}ropagation), a framework for decisive-error localization from step-unlabeled failure trajectories.
At a high level, \soap consists of two components.
\textbf{First}, it constructs a \emph{spectral reference space} from step representations extracted from the failure trajectories.
Each step receives a base error score according to its projection onto a selected band of singular directions.
This produces a step-level signal without requiring decisive-error annotations or a separate collection of verified successful trajectories. \textbf{Second}, \soap addresses downstream contamination through
\emph{dependency-aware rescoring}.
We derive a soft dependency graph from models' attention patterns, where each edge approximates how strongly a later step draws on an earlier one.
Rather than treating downstream failure signals as independent and competing evidence, \soap propagates them backward toward the preceding steps on which they depend.
An earlier step is therefore promoted when anomalous downstream behavior is systematically built upon it.

We evaluate \soap on three failure-attribution benchmarks comprising five subsets---\textsc{Who\&When}, \textsc{Correct-Error}, and \textsc{TraceElephant} using Qwen3.5-9B and DeepSeek-R1-Distill-Llama-8B.
Across different evaluation settings, \soap consistently improves decisive-step localization over competitive prompt-based and training-based baselines.
In particular, on \textsc{TraceElephant} with DeepSeek-R1-Distill-Llama-8B, \soap improves step-level Accuracy by 24.66\% over the strongest baseline.
Ablation studies further show that both spectral step scoring and attention-guided rescoring contribute to the improvements (Section~\ref{sec:exp:analysis}). Our main contributions are summarized as follows:

\vspace{-1em}
\begin{itemize}
    \item We formulate decisive-error localization from
    \emph{step-unlabeled trajectories} within failed trajectories.
    This setting exploits trajectory-level failure signals while requiring neither decisive-error annotations nor carefully verified successful trajectories.

    \item We identify \emph{downstream contamination} as a central obstacle to decisive-error localization, i.e., later steps may exhibit stronger failure signals because they inherit an earlier mistake, causing independent per-step scorers to systematically localize failures too late.

    \item We propose \soap, an effective framework performing spectral step scoring with attention-derived dependency propagation.
    Extensive experiments and ablations on diverse benchmarks demonstrate when and why \soap improves attribution.
\end{itemize}

\section{Problem Setup}
\label{sec:setup}

We formalize the task of localizing the decisive error in a failed agent
trajectory.

\noindent\textbf{Agent trajectory.}
Let $x$ denote a task description and
$\traj=(s_1,\ldots,s_T)$ the resulting trajectory of a multi-agent system.
Each step $s_t$ records the action or message produced at turn $t$ and is
associated with an agent $a_t\in\{1,\ldots,K\}$.
For evaluation, each failed trajectory is paired with a gold decisive-error
index $t^\star\in\{1,\ldots,T\}$, defined as the earliest step at which the
error responsible for the eventual task failure is introduced~\citep{whoandwhen}.
The corresponding responsible agent is $a_{t^\star}$.

Our framework extracts model-internal signals, including hidden
representations, using a frozen language model
$\proxy$.
When the language model underlying the agents exposes its internal states,
$\proxy$ may coincide with that model.
Otherwise, $\proxy$ can be a separate open-weight model, enabling the
analysis of trajectories generated by proprietary agents whose internal
states are inaccessible.
In either case, $\proxy$ is used without fine-tuning on failure-attribution
data.

\noindent \textbf{Failure attribution.}
Given a task description $x$ and a failed trajectory $\traj$, the goal is to
estimate its decisive-error index $t^\star$.
We operationalize this prediction by assigning each step a scalar attribution
score $\Stilde(s_t)\in\mathbb{R}$, where a larger value indicates stronger
evidence that $s_t$ introduced the failure.
The predicted decisive-error step and responsible agent are
$
    \hat{t}
    =
    \arg\max_{t\in\{1,\ldots,T\}}
    \Stilde(s_t),
    \hat{a}
    =
    a_{\hat{t}}.
$
Sorting these scores additionally produces a ranked list of candidate steps
for failure triage.
We evaluate both top-1 localization and top-$k$ performance in
Section~\ref{sec:exp:main} and Appendix~\ref{app:ranked}, respectively.

\noindent \textbf{Step-unlabeled failure trajectories.}
In addition to the test trajectory, we assume access to a disjoint reference
collection of failed trajectories
$
    \mathcal{D}_{\mathrm{fail}}
    =
    \left\{
        \bigl(x^{(n)},\traj^{(n)}\bigr)
    \right\}_{n=1}^{N},
    \traj^{(n)}
    =
    \bigl(s^{(n)}_1,\ldots,s^{(n)}_{T_n}\bigr).
$
The task-level outcome of every reference trajectory is known to be a
failure, but no step-level annotation identifies where the failure was
introduced or which agent was responsible.
We model the  marginal distribution of reference steps as an unlabeled
contamination mixture~\citep{huber1964robust}:
\begin{equation}
\small
    P_{\mathrm{unlabeled}}
    =
    (1-\phi)P_{\mathrm{ord}}
    +
    \phi P_{\mathrm{err}},
    \label{eq:step-contamination}
\end{equation}
where $P_{\mathrm{ord}}$ denotes the distribution of ordinary steps,
$P_{\mathrm{err}}$ denotes that of failure-related steps, and
$\phi\in(0,1)$ is the unknown contamination proportion.
Failure-related steps include the decisive error and any downstream steps
affected by it. Formally, each reference trajectory has an unobserved nonempty set
$\mathcal{E}^{(n)}\subseteq\{1,\ldots,T_n\}$ of failure-related steps, whose
earliest member is the decisive-error step:
$
    t^{\star(n)}
    =
    \min\mathcal{E}^{(n)}.
$
At the marginal level,
$s_t^{(n)}\sim P_{\mathrm{err}}$ when
$t\in\mathcal{E}^{(n)}$ and
$s_t^{(n)}\sim P_{\mathrm{ord}}$ otherwise.
This setting is \textit{practical} because failed trajectories can often be collected
automatically using task-level evaluators, whereas localizing the decisive
error and assigning responsibility require costly expert annotation.

Our method uses $\mathcal{D}_{\mathrm{fail}}$ only to construct the reference
representation space, without observing $\mathcal{E}^{(n)}$ or any
responsible-agent annotations.
Gold step-level labels are reserved for evaluation and, where applicable,
hyperparameter selection on a disjoint validation split.

\begin{figure}[t]
\centering
\includegraphics[width=\textwidth]{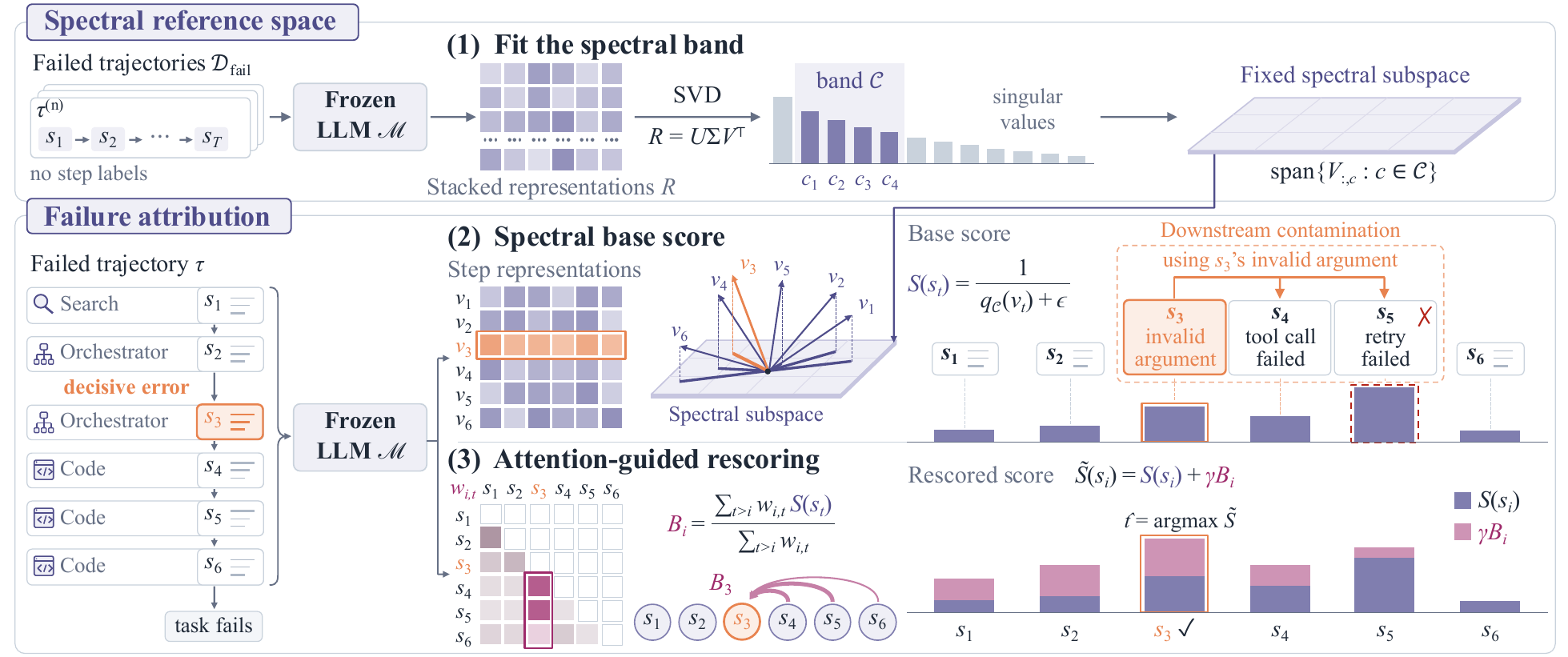}
  \vspace{-0.6cm}
  \caption{\small 
  \textbf{Overview of \soap for agentic failure attribution.} \soap constructs a spectral reference space from step-unlabeled failure trajectories and assigns each step a base error score. To address downstream contamination, \soap derives soft dependencies from attention patterns and propagates downstream error evidence toward likely upstream sources to localize the decisive error.
  }
\label{fig:overview}
\end{figure}


\section{\soap: Spectral Scoring with Attention-Guided Propagation}
\label{sec:method}

We introduce \soap, a framework for localizing decisive errors from
step-unlabeled failure trajectories.
At a high level, \soap consists of two stages.
\textit{First}, we construct a spectral reference space from step
representations extracted by $\proxy$ on $\mathcal{D}_{\mathrm{fail}}$ and
derive a base error score for each step
(\S\ref{sec:method:base}).
\textit{Second}, we estimate soft inter-step dependencies from the attention
patterns of $\proxy$ and propagate error evidence from anomalous downstream
steps back to their likely upstream sources
(\S\ref{sec:method:weights}).
The resulting score $\Stilde(s_t)$ measures how likely step $s_t$ is to have
introduced the failure.
Figure~\ref{fig:overview} summarizes the framework.

\subsection{Spectral Step Scoring from Unlabeled Failure Trajectories}
\label{sec:method:base}

Although every trajectory in $\mathcal{D}_{\mathrm{fail}}$ ends in failure,
most individual steps remain locally valid, while failure-related steps
constitute only a small fraction of each trajectory~\citep{whoandwhen}.
Consequently, the representation space induced by $\proxy$ is dominated by
recurring patterns of ordinary agent behavior, with failure-related steps
forming a sparse, unlabeled component.
We exploit this imbalance by spectrally decomposing the  reference
representations and measuring how strongly each step aligns with selected
spectral directions.

For a trajectory $\traj=(s_1,\ldots,s_T)$, we encode each step using the task
description $x$ and selected trajectory context associated with that step.
Let
$
    v_t\in\mathbb{R}^{d}
$
denote the resulting representation extracted by $\proxy$ for step $s_t$ (cf. Appendix~\ref{app:implementation}).

\noindent \textbf{Spectral reference space.}
We apply the representation extraction procedure to every step in
$\mathcal{D}_{\mathrm{fail}}$, obtaining
$v_t^{(n)}\in\mathbb{R}^{d}$ for the $t$-th step of the $n$-th reference
trajectory.
Let
$
    M_{\mathrm{ref}}
    =
    \sum_{n=1}^{N}T_n
$
denote the total number of reference steps.
Stacking these representations row-wise gives
$R\in\mathbb{R}^{M_{\mathrm{ref}}\times d}$.
Let $r=\min\{M_{\mathrm{ref}},d\}$.
We compute the compact singular value decomposition
\begin{equation}
\small
    R
    =
    U\Sigma V^\top,
    \qquad
    V\in\mathbb{R}^{d\times r}.
    \label{eq:reference-svd}
\end{equation}
The columns of $V$ define orthogonal directions in the reference
representation space, ordered by non-increasing singular value.

Because ordinary steps dominate $R$, the leading singular directions may
primarily encode frequent shared structure, such as response style,
trajectory format, and common tool-use patterns.
Conversely, directions at the extreme tail may capture irrelevant
variation.
We therefore do not restrict scoring to either end of the spectrum.
Instead, we select a contiguous spectral band
\begin{equation}
\small
    \mathcal{C}
    =
    \{c_{\mathrm{begin}},\ldots,c_{\mathrm{end}}\},
    \qquad
    1
    \leq
    c_{\mathrm{begin}}
    \leq
    c_{\mathrm{end}}
    \leq
    r,
    \label{eq:spectral-band}
\end{equation}
whose boundaries are selected as described in
\S\ref{app:implementation}.

\begin{wrapfigure}{r}{0.35\linewidth}
\vspace{-0.9\baselineskip}
\centering
\includegraphics[width=\linewidth]{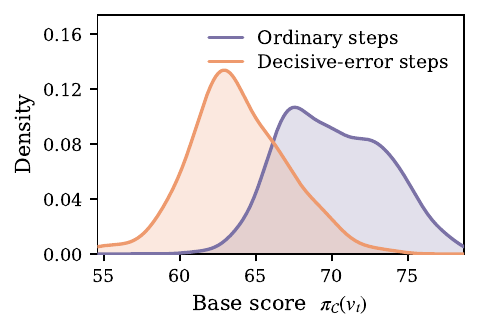}
\vspace{-0.8cm}
\caption{\small \textbf{Base-score distributions} on a \textsc{Correct-Error} test
split~\citep{correct}.
Decisive-error steps exhibit lower projection energy than ordinary steps.}
\label{fig:score-dist}
\vspace{-\baselineskip}
\end{wrapfigure}

\noindent \textbf{Spectral base score.}
For a step representation $v_t$, we compute its average squared projection
onto the selected spectral band and invert this quantity:
\begin{equation}
\small
    q_{\mathcal{C}}(v_t)
    =
    \frac{1}{|\mathcal{C}|}
    \sum_{c\in\mathcal{C}}
    \left\langle v_t,V_{:,c}\right\rangle^2,\quad
    \Sbase(s_t)
    =
    \frac{1}{q_{\mathcal{C}}(v_t)+\epsilon},
\label{eq:base-score}
\end{equation}
where $\epsilon>0$ ensures numerical stability.
Because the reference spectrum is learned primarily from ordinary steps,
decisive-error steps tend to exhibit lower projection energy within the
selected band
(Figure~\ref{fig:score-dist}).
The reciprocal therefore assigns larger base scores to steps exhibiting
stronger decisive-error evidence.
Constructing $R$ and its spectral decomposition \textit{uses no decisive-error or
responsible-agent annotations.}

\subsection{Dependency-aware Rescoring}
\label{sec:method:weights}

\textbf{Attention-guided step dependencies.} The base score in Eq.~\ref{eq:base-score} evaluates each step independently
and therefore cannot distinguish an error source from its downstream
consequences.
Once a decisive error occurs, subsequent steps may inherit the corrupted
context and exhibit stronger failure signals than the original error.
An earlier step is therefore a more plausible root cause when anomalous
downstream steps depend strongly on it.
We capture this structure through a soft dependency graph over the target
trajectory.

For each step $s_t$, let
$
    \mathcal{P}_t
    \subseteq
    \{1,\ldots,t-1\}
$
denote the index set of eligible predecessor steps retained by our context
construction.
For each $i\in\mathcal{P}_t$, we aggregate the post-softmax attention routed
from tokens in $s_t$ to tokens in $s_i$, yielding a non-negative attention
mass $m_{i,t}$.
We normalize these masses over the retained predecessors:
\begin{equation}
\small
    w_{i,t}
    =
    \begin{cases}
        \displaystyle
        \frac{m_{i,t}}
             {\sum_{j\in\mathcal{P}_t}m_{j,t}},
        & i\in\mathcal{P}_t,\\[1.2ex]
        0,
        & \text{otherwise}.
    \end{cases}
    \label{eq:dependency-weight}
\end{equation}
Thus,
$
    \sum_{i\in\mathcal{P}_t}w_{i,t}=1
$
whenever $\mathcal{P}_t$ is nonempty, and $w_{i,t}$ represents the relative
dependency of the later step $s_t$ on the earlier step $s_i$.
The construction of $\mathcal{P}_t$ and the attention aggregation procedure
are detailed in Appendix~\ref{app:implementation}.

These weights provide model-internal evidence about which earlier steps
contribute to the representation of a later step.
Because they are obtained directly from the frozen LLM model, their
construction requires neither repeated generative prompting, which
introduces substantial inference cost~\citep{whoandwhen,oat}, nor explicit
dependency reconstruction and traversal through LLM-based graph
reasoning~\citep{falat,chief,bakish2026adaptive}, nor task-specific
post-training on attributed failure trajectories
~\citep{agentracer,graphtracer}.

\noindent \textbf{Rescoring.} We treat the error signal of a downstream step as additional evidence for
the earlier steps on which it depends.
For each step $s_i$, we compute the dependency-weighted average of its downstream base
scores:
\begin{equation}
\small
    B_i
    =
        \frac{
            \sum_{t>i}
            w_{i,t}\,\Sbase(s_t)
        }{
           \sum_{t>i}w_{i,t}
        },
    \label{eq:backward-evidence}
\end{equation}
In particular, $B_T=0$ because the final step has no downstream dependents. We combine the local error signal with the downstream evidence attributed to
the step:
\begin{equation}
\small
    \Stilde(s_i)
    =
    \Sbase(s_i)
    +
    \gamma B_i,
    \qquad
    \gamma\in[0,1].
    \label{eq:recap}
\end{equation}
Here, $\gamma$ controls the strength of dependency-aware rescoring, with
$\gamma=0$ recovering the spectral base score. The normalization in Eq.~\ref{eq:backward-evidence} prevents an early step
from receiving a large correction merely because it has many downstream
dependents.
Instead, downstream steps with larger dependency weights exert greater
influence on its correction.
Thus, an earlier step is promoted when the downstream steps that depend
strongly on it exhibit large base error scores.

Finally, we predict the decisive-error step and corresponding responsible
agent as
\begin{equation}
\small
    \hat{t}
    =
    \arg\max_{i\in\{1,\ldots,T\}}
    \Stilde(s_i),
    \qquad
    \hat{a}
    =
    a_{\hat{t}}.
    \label{eq:prediction}
\end{equation}
\section{Experiments}
\label{sec:experiments}

\subsection{Setup}
\label{sec:exp:setup}

\textbf{Datasets.} We evaluate our method \soap on 3 failure-attribution benchmarks comprising 5 subsets: 
\textsc{Who\&When}~\citep{whoandwhen}, with algorithm-generated (\textsc{WW-AG}, 126 trajectories) and hand-crafted (\textsc{WW-HC}, 58 trajectories) subsets;
\textsc{Correct-Error}~\citep{correct} (\textsc{CE}, 2{,}226 trajectories);
and \textsc{TraceElephant}~\citep{traceelephant}, with Captain-Agent (\textsc{TE-Cap}, 85 trajectories) and Magentic-One (\textsc{TE-Mag}, 91 trajectories) subsets.
They cover trajectories of varying lengths and diverse QA, coding, and tool-use tasks. Each trajectory is annotated with the decisive-error step and responsible agent. For each subset, we randomly split trajectories into unlabeled-reference, validation, and test sets using a $30/20/50\%$ ratio. The reference annotations are hidden and used only to construct the spectral matrix $R$; validation labels are used for hyperparameter selection, and test labels only for final evaluation. Splitting is performed at the trajectory level. Full statistics and preprocessing details are provided in Appendix~\ref{app:datasets}.

\noindent \textbf{Baselines.} We compare \soap against two families of baselines. First, \emph{prompt-based} methods include All-at-Once, Step-by-Step, and Binary Search~\citep{whoandwhen}, the schema-retrieval method CORRECT~\citep{correct}, the causal-graph method CHIEF~\citep{chief}, and the analyzer--verifier diagnoser ErrorProbe~\citep{errorprobe}. These methods ask a judge LLM to name the decisive step. Second, \emph{training-based} methods include OAT~\citep{oat}, which learns successful-trajectory dynamics through one-class training, and StepFinder~\citep{zhu2026stepfinder}, which trains a step classifier on synthetically corrupted trajectories with step-level error labels.

\noindent \textbf{Implementation details.} We use Qwen3.5-9B~\citep{qwen3} and DeepSeek-R1-Distill-Llama-8B~\citep{deepseekr1} (DeepSeek-8B) as backbones $\proxy$. We obtain step representations by mean-pooling token hidden states. We select the representation layer, the spectral band $\mathcal{C}$, the attention layer band, and the propagation strength $\gamma$ on the validation set. All training-based baselines use the same two backbones and test data as \soap, with the hyperparameters recommended in their papers. Following~\citet{oat}, we additionally compare with prompt-based baselines with GPT-4o as the judge (results with GPT-5 are in Appendix~\ref{app:prompting-gpt5}). Following \citet{whoandwhen}, we report \emph{Step-Level Accuracy} (additional metrics, including \textit{Agent-level Accuracy} and \textit{top-$k$ Accuracy} are in Appendix~\ref{app:metrics} and~\ref{app:ranked}). Further implementation details are provided in Appendices~\ref{app:implementation} and~\ref{app:baselines}.

\begin{table}[!t]
\centering
\setlength{\tabcolsep}{6pt}
\renewcommand{\arraystretch}{1.04}
\caption{\small \textbf{Main comparison in the \emph{without-ground-truth} setting.} Results are step-level attribution accuracy (\%), averaged over three seeds (std in Appendix~\ref{app:metrics}). No method receives the gold task answer during inference. \textbf{Supervised} marks methods that consume step-level error annotations. Best per column within each block in {bold}, second best \second{underlined}. } 
\vspace{-0.3cm}
\label{tab:main}
\begin{tabularx}{\textwidth}{@{}lc *{5}{>{\centering\arraybackslash}X}@{}}
\toprule
Method & Supervised & \textsc{WW-AG} & \textsc{WW-HC} & \textsc{CE} & \textsc{TE-Cap} & \textsc{TE-Mag} \\
\midrule
\rowcolor{Gray}
\multicolumn{7}{c}{\textit{\textbf{GPT-4o}}}\\
\multicolumn{7}{@{}l}{\textit{Prompt-based methods}}\\
\quad All-at-Once~\citep{whoandwhen}   & \xmark & 14.81 & 6.90 & 28.39 & \second{16.28} & 5.80 \\
\quad Step-by-Step~\citep{whoandwhen}  & \xmark & 20.63 & 13.79 & \second{44.47} & 13.95 & 13.04 \\
\quad Binary Search~\citep{whoandwhen} & \xmark & 22.22 & 4.60 & 9.70 & 9.30 & 6.52 \\
\quad CORRECT~\citep{correct}          & \cmark & 15.34 & 4.60 & 35.12 & 10.85 & \second{13.77} \\
\quad CHIEF~\citep{chief}              & \xmark & \second{25.93} & \second{14.94} & 38.56 & 13.95 & 8.70 \\
\quad ErrorProbe~\citep{errorprobe}    & \xmark & \textcolor{gray}{\best{38.62}} & \textcolor{gray}{\best{21.84}} & \textcolor{gray}{\best{59.52}} & \textcolor{gray}{\best{27.13}} & \textcolor{gray}{\best{28.26}} \\
\midrule
\rowcolor{Gray}
\multicolumn{7}{c}{\textit{\textbf{Qwen3.5-9B}}}\\
\multicolumn{7}{@{}l}{\textit{Prompt-based methods}}\\
\quad All-at-Once~\citep{whoandwhen}    &    \xmark    & 17.46 & 11.49 & 33.47 & 4.65 & 1.45 \\
\quad Step-by-Step~\citep{whoandwhen}   &    \xmark    & 10.58 & 16.09 & 31.15 & 14.73 & 10.87 \\
\quad Binary Search~\citep{whoandwhen}  &    \xmark    & 2.65 & 13.79 & \second{59.92} & 0.78 & 5.07 \\
\quad CORRECT~\citep{correct}       & \cmark & 29.10 & 3.45 & 36.35 & 7.75 & 2.90 \\
\quad CHIEF~\citep{chief}         &     \xmark   & 28.57 & 3.45 & 29.39 & 3.88 & 2.17 \\
\quad ErrorProbe~\citep{errorprobe}    &    \xmark    & \second{39.68} & \second{21.84} & 55.03 & \second{23.26} & \second{20.29} \\
\addlinespace[1pt]
\multicolumn{7}{@{}l}{\textit{Training-based methods}}\\
\quad StepFinder~\citep{zhu2026stepfinder} & \cmark & 15.87 & 13.33 & 30.03 & 18.29 & 6.67 \\
\quad OAT~\citep{oat}                  & \xmark & 16.72 & 10.11 & 56.50 & 15.35 & 18.84 \\
\addlinespace[1pt]\midrule
\textbf{\soap (ours)}            & \xmark & \best{47.62} & \best{34.48} & \best{61.78} & \best{35.66} & \best{23.19} \\
\midrule
\rowcolor{Gray}
\multicolumn{7}{c}{\textit{\textbf{DeepSeek-R1-Distill-Llama-8B}}}\\
\multicolumn{7}{@{}l}{\textit{Prompt-based methods}}\\
\quad All-at-Once~\citep{whoandwhen}   &    \xmark    & 7.41 & 5.75 & 21.95 & 4.65 & 2.90 \\
\quad Step-by-Step~\citep{whoandwhen}  &    \xmark    & 16.93 & 9.20 & 9.04 & 11.63 & 4.35 \\
\quad Binary Search~\citep{whoandwhen} &   \xmark     & 7.94 & 12.64 & 34.15 & 10.08 & \second{13.77} \\
\quad CORRECT~\citep{correct}       & \cmark & 20.11 & 11.49 & 32.82 & 3.88 & 0.72 \\
\quad CHIEF~\citep{chief}         &     \xmark   & 17.46 & 2.30 & 8.40 & 7.75 & 9.42 \\
\quad ErrorProbe~\citep{errorprobe}    &    \xmark    & \second{24.34} & 10.34 & 35.97 & 17.05 & 8.70 \\
\addlinespace[1pt]
\multicolumn{7}{@{}l}{\textit{Training-based methods}}\\
\quad StepFinder~\citep{zhu2026stepfinder} & \cmark & 18.94 & 10.80 & 36.91 & 14.73 & 4.20 \\
\quad OAT~\citep{oat}                  & \xmark & 12.70 & \second{15.86} & \second{52.50} & \second{17.98} & 13.04 \\
\addlinespace[1pt]\midrule
\textbf{\soap (ours)}            & \xmark & \best{45.50} & \best{28.74} & \best{64.68} & \best{42.64} & \best{30.43} \\
\bottomrule
\end{tabularx}
\end{table}

\subsection{Main Results}
\label{sec:exp:main}
\begin{figure}[t]
\centering
\includegraphics[width=\textwidth]{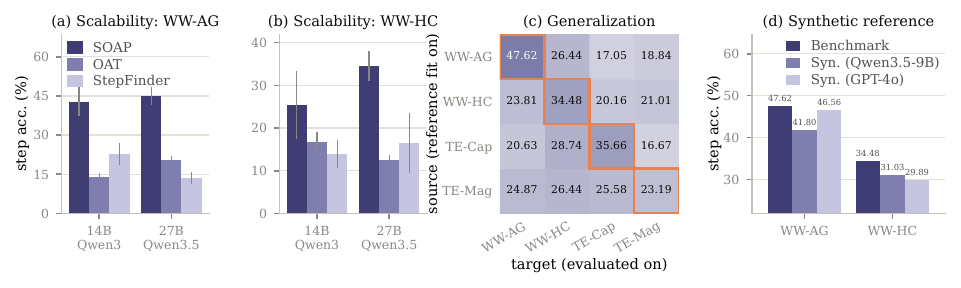}
\vspace{-0.7cm}
\caption{\small (a--b) \textbf{Scalability to larger backbones}, which show the step-level accuracy (\%) on \textsc{WW-AG} and \textsc{WW-HC} using Qwen3-14B and Qwen3.5-27B; error bars denote standard deviations.
{(c)} \textbf{Cross-distribution generalization}, with source datasets as rows and target datasets as columns; outlined diagonal entries correspond to the in-distribution results in Table~\ref{tab:main}.
{(d)} \textbf{Results with synthetic trajectories} generated by Qwen3.5-9B and GPT-4o.
Panels (c) and (d) use Qwen3.5-9B as LLM $\proxy$ for embedding extraction and rescoring. }

\label{fig:scale}
\label{fig:transfer}
\end{figure}

\begin{wraptable}{r}{0.45\textwidth}
\vspace{-0.3\baselineskip}
\centering
\scriptsize
\setlength{\tabcolsep}{5.5pt}
\renewcommand{\arraystretch}{0.98}
\caption{\small \textbf{Comparison in the \emph{with-ground-truth} setting.} Results are step-level accuracy (\%) on \textsc{Who\&When}, with the gold task answer provided. All methods use Qwen3.5-9B as the backbone. Marks follow Table~\ref{tab:main}. }
\vspace{-0.3cm}
\label{tab:main-gt}
\resizebox{\linewidth}{!}{%
\begin{tabular}{@{}lc cc@{}}
    \toprule
    Method & Sup. & \textsc{WW-AG} & \textsc{WW-HC} \\
    \midrule
    \multicolumn{4}{@{}l}{\textit{Prompt-based methods}}\\
    \quad All-at-Once   & \xmark& 21.16 & 11.49 \\
    \quad Step-by-Step  & \xmark& 17.46 & 18.39 \\
    \quad Binary Search & \xmark& 2.12 & 13.79 \\
    \quad CORRECT       & \cmark& 32.28 & 4.60 \\
    \quad CHIEF         & \xmark& 30.69 & 4.60 \\
    \quad ErrorProbe    & \xmark& \second{40.74} & \second{19.54} \\
    \addlinespace[1pt]
    \multicolumn{4}{@{}l}{\textit{Training-based methods}}\\
    \quad StepFinder    & \cmark& 18.52 & 12.87 \\
    \quad OAT           & \xmark& 22.12 & 8.05 \\
    \addlinespace[1pt]\cmidrule(lr){1-4}
    \quad \textbf{\soap (ours)} & \xmark& \best{43.39} & \best{34.48} \\
    \bottomrule
\end{tabular}}
\vspace{-1\baselineskip}
\end{wraptable}

\textbf{Comparison with competitive baselines.} Table~\ref{tab:main} compares \soap with prompt-based and training-based baselines across all five subsets, without access to the gold task answer at inference. With both open-weight backbones, \soap achieves the highest step-level accuracy on every subset. For example, on \textsc{WW-AG} with Qwen3.5-9B, \soap attains 47.62\%, compared with 39.68\% for ErrorProbe~\citep{errorprobe}, the strongest prompt-based baseline, and 16.72\% for OAT~\citep{oat}, the strongest training-based baseline. Its advantage over prompt-based methods using closed-source judges also persists with both GPT-4o and GPT-5; the GPT-5 results are  in Appendix~\ref{app:prompting-gpt5}.  These results demonstrate that step-unlabeled failure trajectories provide an effective source of localization signal, outperforming the success-only trajectories used by OAT and the step-level supervision used by StepFinder. Agent-level results  are provided in Appendix~\ref{app:metrics}.

\noindent \textbf{Results for the with-GT setting.} To assess the effectiveness of \soap under the with-GT setting, we provide the gold task answer to all methods and compare them in Table~\ref{tab:main-gt}. When provided with the ground-truth answer, \soap remains the best method on both subsets, ahead of the strongest prompt-based baseline ErrorProbe~\citep{errorprobe} (43.39 vs.\ 40.74 on \textsc{WW-AG} and 34.48 vs.\ 19.54 on \textsc{WW-HC}), while every training-based baseline trails by more than 20 points (e.g., 22.12 for OAT~\citep{oat} on \textsc{WW-AG}). Results on DeepSeek-8B appear in Appendix~\ref{app:gt}. 

\noindent \textbf{Scalability to larger backbones.} To assess scalability, we evaluate \soap on two larger backbones, Qwen3-14B and Qwen3.5-27B. As shown in Figure~\ref{fig:scale}(a-b), \soap achieves stable accuracy on both backbones and outperforms the SOTA baselines (e.g., 44.97 vs.\ 22.65 for StepFinder~\citep{zhu2026stepfinder} on \textsc{WW-AG}). These results confirm that \soap remains effective on larger backbones.

\noindent \textbf{Generalization across data distributions.} We construct the reference on a source subset and directly evaluate \soap on a different target subset, for every pair among \{\textsc{WW-AG}, \textsc{WW-HC}, \textsc{TE-Cap}, \textsc{TE-Mag}\}. As shown in Figure~\ref{fig:transfer}(c), \soap transfers across subsets: for example, constructing on \textsc{WW-HC} and testing on \textsc{TE-Mag} achieves 21.01, compared with the in-domain result of 23.19. These results indicate that the spectral reference captures failure structure that can be generalizable in practice. Results on DeepSeek-8B appear in Appendix~\ref{app:transfer}.

\noindent \textbf{Results with synthetic trajectories.}  
Our main results construct $R$ from existing benchmark trajectories. We further examine whether alternative synthetic trajectories can serve as the reference while retaining Qwen3.5-9B as the attribution backbone. We generate the same number of trajectories using either Qwen3.5-9B itself or GPT-4o as an external generator, while keeping the original validation and test splits unchanged. As shown in Figure~\ref{fig:transfer}(d), both synthetic references remain effective. For example, the GPT-4o-generated reference achieves 46.56 on \textsc{WW-AG}, close to 47.62 using the original training trajectories. Thus, \soap can construct its reference space from synthetic trajectories, either when generated by the same or a different model. Generation details are provided in Appendix~\ref{app:synth}.

\begin{figure}[t]
\centering
\includegraphics[width=\textwidth]{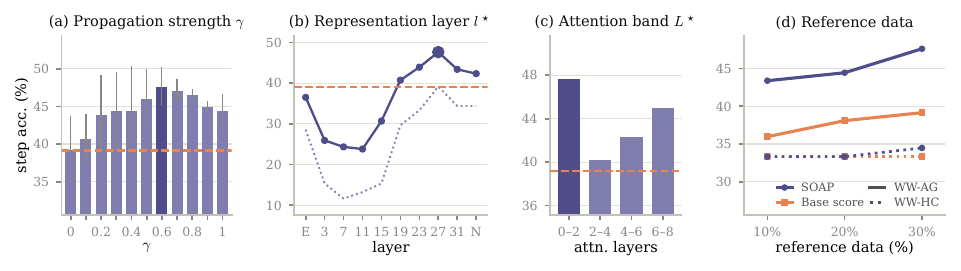}
\vspace{-0.9cm}
\caption{\small \textbf{Ablation studies.} Sensitivity to (a) propagation strength $\gamma$, (b) representation layer $l^\star$, (c) attention layer band $L^\star$, and (d) the amount of unlabeled reference data. All experiments use Qwen3.5-9B. Panels (a--c) vary one hyperparameter at a time from the selected \textsc{WW-AG} configuration used in Table~\ref{tab:main}. Dark bars and enlarged markers denote the selected values; the orange dashed line denotes the base score without rescoring ($\gamma=0$), and the dotted curve in (b) reports the base score at each layer. Panel (d) constructs $R$ using $10\%$, $20\%$, or $30\%$ of the reference corpus and reports results on \textsc{WW-AG} and \textsc{WW-HC}. Best viewed in color. }
\label{fig:datasize}
\end{figure}

\vspace{-1em}

\subsection{Analysis}
\label{sec:exp:analysis}

\begin{table}[t]
\begin{minipage}[t]{0.47\textwidth}
\centering
\footnotesize
\setlength{\tabcolsep}{4pt}
\renewcommand{\arraystretch}{1.04}
\caption{\small \textbf{Comparison with alternative scoring functions.} Results are step-level accuracy (\%) on backbone Qwen3.5-9B, with no rescoring.
}
\vspace{-0.3cm}
\label{tab:scorefn}
\begin{tabularx}{\linewidth}{l*{2}{>{\centering\arraybackslash}X}}
    \toprule
    Scoring function & \textsc{WW-AG} & \textsc{WW-HC} \\
    \midrule
    Perplexity           & 3.70  & 8.05 \\
    Random subspace      & 13.23 & 9.20 \\
    Top subspace         & 12.70 & \best{33.33} \\
    Tail subspace        & 6.35  & 6.90 \\
    Full spectrum        & 12.17 & 31.03 \\
    Norm $\ell_1$        & 18.52 & 18.39 \\
    Norm $\ell_2$        & 14.29 & 26.44 \\
    \rowcolor{Gray}
    \textbf{Spectral band (ours)} & \best{39.15} & \best{33.33} \\
    \bottomrule
    \end{tabularx}
\end{minipage}\hfill
\begin{minipage}[t]{0.49\textwidth}
\centering
\footnotesize
\setlength{\tabcolsep}{4pt}
\renewcommand{\arraystretch}{1.04}
\caption{\small \textbf{Alternatives to attention-guided rescoring.}
Step-level accuracy using Qwen3.5-9B; all methods share the same base score.}
\label{tab:weights}
\vspace{-0.3cm}
\begin{tabularx}{\linewidth}{l*{2}{>{\centering\arraybackslash}X}}
\toprule
Rescoring design & \textsc{WW-AG} & \textsc{CE} \\
\midrule
Base score (no rescoring) & 39.15 & 61.38 \\
\addlinespace[2pt]
\multicolumn{3}{@{}l}{\textit{Content-agnostic}} \\
\quad Uniform, unnormalized & 16.93 & 57.85 \\
\quad Uniform, normalized   & 40.21 & 60.78 \\
\addlinespace[2pt]
\multicolumn{3}{@{}l}{\textit{Position heuristics}} \\
\quad Temporal bias, std. & 40.74 & 61.62 \\
\quad Temporal bias, base      & 22.22 & 41.65 \\
\quad Earliest of top-$5$      & 21.16 & 4.51 \\
\addlinespace[2pt]
\rowcolor{Gray}
\textbf{Attention-guided (ours)} & \best{47.62} & \best{61.78} \\
\bottomrule
\end{tabularx}
\end{minipage}
\end{table}
\textbf{Comparison with alternative scoring functions.} We compare the spectral base score (Eq.~\ref{eq:base-score}) with seven alternatives in Table~\ref{tab:scorefn}: step perplexity, projection to random subspace, top subspace, tail subspace, full spectrum, and the $\ell_1$ and $\ell_2$ norms of LLM step embeddings (details in Appendix~\ref{app:implementation}). Our spectral band (Eq.~\ref{eq:spectral-band}) remains the best on both subsets, and the best alternative reaches only 18.52 on \textsc{WW-AG}, compared with 39.15 for ours. These results indicate that the decisive-error signal lies more likely in a selected spectral band  than in LLM predictive likelihood or embedding magnitude.

\noindent \textbf{Sensitivity to the hyperparameters.} We vary the propagation strength $\gamma$ (Eq.~\ref{eq:recap}), representation layer $l^\star$ for base score calculation (Eq.~\ref{eq:base-score}), and layer band $L^\star$ for attention mass calculation (Eq.~\ref{eq:dependency-weight}) one at a time. Here, $L^\star$ specifies the group of attention blocks averaged to estimate step dependencies (cf. Appendix~\ref{app:implementation} for details). As shown in Figure~\ref{fig:datasize}, propagation consistently improves over the base score and peaks at $\gamma=0.6$, while later representation layers perform best. The earliest attention band performs best, suggesting that lower attention blocks provide clearer dependency signals, although all bands improve over the base score.  Extended ablations on subsets \textsc{WW-HC}, \textsc{TE-Cap}, \textsc{TE-Mag}, backbone DeepSeek-8B are in Appendix~\ref{app:deepseek-ablations}.

\noindent \textbf{Importance of attention-guided rescoring.}
Table~\ref{tab:weights} keeps the spectral base score fixed and replaces attention-guided rescoring with simpler alternatives (details in Appendix~\ref{app:implementation}). \emph{Uniform, normalized} averages all successor scores equally, whereas \emph{uniform, unnormalized} sums them, favoring early steps with more successors. \emph{Temporal bias} subtracts a penalty $\lambda t/T$ that grows with the step position $t$, either from the base score $\Sbase(s_t)$ directly (\emph{base}) or after standardizing the base scores within each trajectory (\emph{standardized}). \emph{Earliest of top-$5$} selects the earliest among the five highest-scoring steps. Attention-guided rescoring performs best on both subsets, reaching 47.62 on \textsc{WW-AG} compared with 40.74 for the strongest alternative. This shows that the gain arises from selectively weighting downstream evidence through attention, rather than uniform aggregation or positional bias.

\begin{wrapfigure}{r}{0.40\textwidth}
\vspace{-0.9\baselineskip}
\centering
\includegraphics[width=0.95\linewidth]{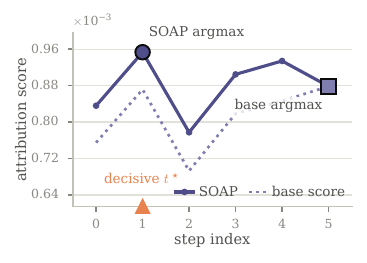}
\vspace{-0.3cm}
\caption{\textbf{Qualitative example.} Per-step scores on a \textsc{TE-Cap} trajectory with backbone Qwen3.5-9B. The base score (dotted) peaks late; \soap (solid) lifts the decisive error $t^\star$ (orange marker) to the top. The large square and circle mark each score's argmax.}
\label{fig:qualitative}
\vspace{-0.9\baselineskip}
\end{wrapfigure}

\noindent \textbf{Quantity of unlabeled reference data.} We construct $R$ on random subsamples covering $\{10, 20, 30\}\%$ of the corpus, with the validation and test splits fixed. As shown in Figure~\ref{fig:datasize}(d), \soap needs little reference data: with a third of our 30\% train split (twelve trajectories on \textsc{WW-AG} and six on \textsc{WW-HC}), accuracy is within a few points of the full-data result (43.39 vs.\ 47.62 on \textsc{WW-AG}). These results indicate that a handful of unlabeled trajectories suffices to construct the reference, showcasing the practicality of our approach.

\noindent \textbf{Qualitative example.} In the \textsc{TE-Cap} trajectory of Figure~\ref{fig:qualitative}, the base score peaks on the final Orchestrator turn, which only restates the wrong answer. Rescoring routes that late signal back to the decisive error four steps earlier, where the extraction expert misreads the program's structure and proposes the wrong fix, and corrects the prediction. More textual examples are in Appendix~\ref{app:qualitative}. 

\noindent \textbf{Additional analysis.} Due to space limitations, we defer additional experiments to the Appendix: cost analysis~(\ref{app:compute}), context window $w$ in constructing $\mathcal{P}_t$~(\ref{app:window}), representation pooling~(\ref{app:pooling}), hyperparameter sensitivity~(\ref{app:sensitivity}), trivial baselines~(\ref{app:length}), effect of the reference corpus~(\ref{app:reference-controls}), a supervised probe comparison~(\ref{app:probe}), and extended results on larger backbones~(\ref{app:proxies}), synthetic references~(\ref{app:synth-results}), and reference data quantity~(\ref{app:datasize}).

\vspace{-1em}
\section{Related Work}
\label{sec:related}

\textbf{Agentic failure attribution} has gained interest recently for ensuring LLM-based agents' reliability~\citep{qi2026trajdebug,barke2026agentrx}.
\citet{whoandwhen} formulate automated failure attribution, introduce the
\textsc{Who\&When} benchmark, and evaluate All-at-Once, Step-by-Step, and Binary Search
LLM judges; Who\&When~Pro scales the benchmark to 12K failed trajectories
across modalities~\citep{liu2026wwpro}, and complementary work categorizes
common multi-agent failure
modes~\citep{mast,deshpande2025trail,raj2026model,zhu2025agentdebug}.
Subsequent methods improve inference-time prompting: RAFFLES iterates a
judge--evaluator loop~\citep{raffles}, ErrorProbe pairs an analyzer with a
self-improving verifier~\citep{errorprobe}, CHIEF constructs causal dependency
graphs through multi-stage LLM reasoning~\citep{chief}, and others
restructure the context, retrieve error schemata, or reconstruct step
dependencies~\citep{echo,correct,bakish2026adaptive,falat}. Another line
trains dedicated attributors: AgenTracer and StepFinder learn from labeled or
synthetically corrupted trajectories by reinforcement learning or
temporal-semantic modeling~\citep{agentracer,zhu2026stepfinder}, and
GraphTracer learns to trace root causes over information-flow
graphs~\citep{graphtracer}. These approaches, however, rely on
\textit{computationally intensive inference or step-level error annotations},
limiting their practicality; we compare with representatives of both lines in
Table~\ref{tab:main}. OAT avoids both by one-class learning over clean
successful trajectories~\citep{oat}, but its success-only data \textit{carry
no decisive-error signal}, unlike our step-unlabeled failure trajectories,
and it underperforms \soap (Table~\ref{tab:main}).  \citet{du2024haloscope} detects hallucinations using unlabeled LLM generations but trains a classifier on top of a different projection score without the dependency propagation compared to our approach. \citet{masprism} also use
attention signals, but score them on steps with high negative log-likelihood
for two-stage localization; MASC targets online self-correction via
prototype-guided next-execution reconstruction rather than post-hoc
error attribution~\citep{shen2025masc}.

\vspace{-1em}
\section{Conclusion}
\label{sec:conclusion}

We propose \soap, an effective framework for failure attribution in LLM-based multi-agent systems. Our framework scores each step against a spectral reference constructed on unlabeled failure trajectories, and routes late contamination  back to the decisive error with attention-guided propagation. \soap is lightweight, requires no step-level annotation, and can be readily applied to off-the-shelf LLMs. Extensive experiments show that \soap consistently outperforms prompt-based and training-based baselines across benchmarks, backbones, and settings. We hope our work can inspire future research on failure attribution and more broadly on building reliable multi-agent systems.

\newpage
\bibliography{iclr2026_conference}

\begin{thebibliography}{29}
\providecommand{\natexlab}[1]{#1}
\providecommand{\url}[1]{\texttt{#1}}
\expandafter\ifx\csname urlstyle\endcsname\relax
  \providecommand{\doi}[1]{doi: #1}\else
  \providecommand{\doi}{doi: \begingroup \urlstyle{rm}\Url}\fi

\bibitem[Bakish et~al.(2026)Bakish, Dudai, Ganz, Nuriel, Ben~Avraham, Nacson, and Litman]{bakish2026adaptive}
Y.~Bakish, A.~Dudai, R.~Ganz, O.~Nuriel, E.~Ben~Avraham, M.~S. Nacson, and R.~Litman.
\newblock Adaptive influence graphs for failure attribution in multi-agent systems.
\newblock \emph{arXiv preprint arXiv:2608.24361}, 2026.

\bibitem[Bandi et~al.(2026)Bandi, Dumitru, Hertzberg, Agarwal, Boo, Polakam, Hassaan, Da, Kim, Gupta, Sharma, Park, Dimakis, Hernandez~Montoya, Rambado, Salazar, Cruz, Rezaei, Rane, Levin, Zhang, Kenstler, and Liu]{bandi2026mcpatlas}
C.~Bandi, R.-G. Dumitru, B.~Hertzberg, D.~Agarwal, G.~Boo, T.~Polakam, S.~Hassaan, J.~Da, H.~Kim, V.~Gupta, M.~Sharma, A.~Park, M.~Dimakis, E.~G. Hernandez~Montoya, D.~Rambado, I.~Salazar, R.~Cruz, M.~Rezaei, C.~Rane, B.~Levin, D.~Y. Zhang, B.~Kenstler, and B.~Liu.
\newblock {MCP-Atlas}: A large-scale benchmark for tool-use competency with real {MCP} servers.
\newblock \emph{arXiv preprint arXiv:2602.00933}, 2026.

\bibitem[Banerjee et~al.(2025)]{echo}
A.~Banerjee et~al.
\newblock Where did it all go wrong? a hierarchical look into multi-agent error attribution.
\newblock \emph{arXiv preprint arXiv:2510.04886}, 2025.

\bibitem[Barke et~al.(2026)Barke, Goyal, Khare, Singh, Nath, and Bansal]{barke2026agentrx}
S.~Barke, A.~Goyal, A.~Khare, A.~Singh, S.~Nath, and C.~Bansal.
\newblock {AgentRx}: Diagnosing {AI} agent failures from execution trajectories.
\newblock \emph{arXiv preprint arXiv:2602.02475}, 2026.

\bibitem[Cemri et~al.(2026)Cemri, Pan, Yang, Agrawal, Chopra, Tiwari, Keutzer, Parameswaran, Klein, Ramchandran, Zaharia, Gonzalez, and Stoica]{mast}
M.~Cemri, M.~Z. Pan, S.~Yang, L.~A. Agrawal, B.~Chopra, R.~Tiwari, K.~Keutzer, A.~Parameswaran, D.~Klein, K.~Ramchandran, M.~Zaharia, J.~E. Gonzalez, and I.~Stoica.
\newblock Why do multi-agent {LLM} systems fail?
\newblock In \emph{The Thirty-ninth Annual Conference on Neural Information Processing Systems Datasets and Benchmarks Track}, 2026.
\newblock URL \url{https://openreview.net/forum?id=fAjbYBmonr}.

\bibitem[Chen et~al.(2026)Chen, Wang, Mu, Wang, Liu, Feng, and Wang]{traceelephant}
M.~Chen, J.~Wang, F.~Mu, Y.~Wang, Z.~Liu, H.~Feng, and Q.~Wang.
\newblock Seeing the whole elephant: A benchmark for failure attribution in {LLM}-based multi-agent systems.
\newblock In M.~Liakata, V.~P. Moreira, J.~Zhang, and D.~Jurgens, editors, \emph{Proceedings of the 64th Annual Meeting of the {A}ssociation for {C}omputational {L}inguistics (Volume 1: Long Papers)}, pages 19888--19905, San Diego, California, United States, July 2026. Association for Computational Linguistics.
\newblock ISBN 979-8-89176-390-6.
\newblock \doi{10.18653/v1/2026.acl-long.912}.
\newblock URL \url{https://aclanthology.org/2026.acl-long.912/}.

\bibitem[Deshpande et~al.(2025)Deshpande, Gangal, Mehta, Krishnan, Kannappan, and Qian]{deshpande2025trail}
D.~Deshpande, V.~Gangal, H.~Mehta, J.~Krishnan, A.~Kannappan, and R.~Qian.
\newblock Trail: Trace reasoning and agentic issue localization.
\newblock \emph{arXiv preprint arXiv:2505.08638}, 2025.

\bibitem[Du et~al.(2024)Du, Xiao, and Li]{du2024haloscope}
X.~Du, C.~Xiao, and Y.~Li.
\newblock Haloscope: Harnessing unlabeled llm generations for hallucination detection.
\newblock \emph{Advances in Neural Information Processing Systems}, 37:\penalty0 102948--102972, 2024.

\bibitem[Guo et~al.(2025)Guo, Yang, Zhang, Song, Zhang, Xu, et~al.]{deepseekr1}
D.~Guo, D.~Yang, H.~Zhang, J.~Song, R.~Zhang, R.~Xu, et~al.
\newblock Deepseek-r1: Incentivizing reasoning capability in llms via reinforcement learning.
\newblock \emph{arXiv preprint arXiv:2501.12948}, 2025.

\bibitem[Huber(1964)]{huber1964robust}
P.~J. Huber.
\newblock Robust estimation of a location parameter.
\newblock \emph{The Annals of Mathematical Statistics}, 35\penalty0 (1):\penalty0 73--101, 1964.
\newblock \doi{10.1214/aoms/1177703732}.

\bibitem[Li et~al.(2026)Li, Yilmaz, Chen, and Le]{errorprobe}
J.~Li, E.~Yilmaz, B.~Chen, and T.~Le.
\newblock Towards self-improving error diagnosis in multi-agent systems.
\newblock In M.~Liakata, V.~P. Moreira, J.~Zhang, and D.~Jurgens, editors, \emph{Findings of the {A}ssociation for {C}omputational {L}inguistics: {ACL} 2026}, pages 2063--2077, San Diego, California, United States, July 2026. Association for Computational Linguistics.
\newblock ISBN 979-8-89176-395-1.
\newblock \doi{10.18653/v1/2026.findings-acl.98}.
\newblock URL \url{https://aclanthology.org/2026.findings-acl.98/}.

\bibitem[Liu et~al.(2026{\natexlab{a}})Liu, Xi, Zhang, Zeng, Yue, Wang, Kang, Wu, and Wang]{liu2026wwpro}
J.~Liu, H.~Xi, S.~Zhang, Y.~Zeng, T.~Yue, C.~Wang, J.~Kang, Q.~Wu, and H.~Wang.
\newblock Who\&when pro: Can {LLMs} really attribute failures in {AI} agents?
\newblock \emph{arXiv preprint arXiv:2607.09996}, 2026{\natexlab{a}}.

\bibitem[Liu et~al.(2026{\natexlab{b}})Liu, Feng, Pu, and Chen]{masprism}
Y.~Liu, H.~Feng, J.~Pu, and Z.~Chen.
\newblock {MASPrism}: Lightweight failure attribution for multi-agent systems using prefill-stage signals.
\newblock \emph{arXiv preprint arXiv:2605.07509}, 2026{\natexlab{b}}.

\bibitem[Qi et~al.(2026)Qi, Yin, Shi, Peng, Lu, Liu, Xuan, Liu, Hu, Wang, Hou, Xu, and Li]{qi2026trajdebug}
Y.~Qi, Z.~Yin, X.~Shi, H.~Peng, S.~Lu, Y.~Liu, R.~Xuan, Y.~Liu, Z.~Hu, X.~Wang, L.~Hou, B.~Xu, and J.~Li.
\newblock {TRAJDEBUG}: Tracing error lifecycle to identify critical failures in long-horizon agent trajectories.
\newblock \emph{arXiv preprint arXiv:2608.06346}, 2026.

\bibitem[{Qwen Team}(2025)]{qwen3}
{Qwen Team}.
\newblock Qwen3 technical report.
\newblock \emph{arXiv preprint arXiv:2505.09388}, 2025.

\bibitem[Rafi et~al.(2026)Rafi, Ahasanuzzaman, Kim, Wang, and Chen]{falat}
M.~N. Rafi, M.~Ahasanuzzaman, D.~J. Kim, Z.~Wang, and T.-H. Chen.
\newblock Falat: Tracing failures in llm agent trajectories via dependency-guided search, 2026.
\newblock URL \url{https://arxiv.org/abs/2606.00765}.

\bibitem[Raj et~al.(2026)Raj, Gupta, Mahmoud, Dumitru, Yi, Sabharwal, and He]{raj2026model}
H.~Raj, V.~Gupta, A.~Mahmoud, R.-G. Dumitru, D.~Yi, A.~Sabharwal, and Y.~He.
\newblock Model or harness? an interaction-centric taxonomy for localizing agent failures.
\newblock \emph{arXiv preprint arXiv:2607.28802}, 2026.

\bibitem[Shen et~al.(2025)Shen, Zhang, Wang, Tan, Zhao, Yao, Tadiparthi, Mahjoub, Pari, Lee, and Chen]{shen2025masc}
X.~Shen, Q.~Zhang, S.~Wang, Z.~Tan, X.~Zhao, L.~Yao, V.~Tadiparthi, H.~N. Mahjoub, E.~M. Pari, K.~Lee, and T.~Chen.
\newblock Metacognitive self-correction for multi-agent system via prototype-guided next-execution reconstruction.
\newblock \emph{arXiv preprint arXiv:2510.14319}, 2025.

\bibitem[Wang et~al.(2024)Wang, Ma, Feng, Zhang, Yang, Zhang, Chen, Tang, Chen, Lin, Zhao, Wei, and Wen]{wang2024surveyagents}
L.~Wang, C.~Ma, X.~Feng, Z.~Zhang, H.~Yang, J.~Zhang, Z.~Chen, J.~Tang, X.~Chen, Y.~Lin, W.~X. Zhao, Z.~Wei, and J.~Wen.
\newblock A survey on large language model based autonomous agents.
\newblock \emph{Frontiers of Computer Science}, 18\penalty0 (6):\penalty0 186345, 2024.
\newblock \doi{10.1007/s11704-024-40231-1}.

\bibitem[Wang et~al.(2026)Wang, Wu, Wang, and Wang]{chief}
Y.~Wang, W.~Wu, J.~Wang, and Q.~Wang.
\newblock From flat logs to causal graphs: Hierarchical failure attribution for llm-based multi-agent systems, 2026.
\newblock URL \url{https://arxiv.org/abs/2602.23701}.

\bibitem[Yao et~al.(2023)Yao, Zhao, Yu, Du, Shafran, Narasimhan, and Cao]{yao2023react}
S.~Yao, J.~Zhao, D.~Yu, N.~Du, I.~Shafran, K.~R. Narasimhan, and Y.~Cao.
\newblock {ReAct}: Synergizing reasoning and acting in language models.
\newblock In \emph{International Conference on Learning Representations}, 2023.
\newblock URL \url{https://openreview.net/forum?id=WE_vluYUL-X}.

\bibitem[Yeh et~al.(2026)Yeh, Zhu, Deep, and Li]{oat}
S.~Yeh, Y.~Zhu, S.~Deep, and S.~Li.
\newblock Tracing agentic failure from the flow of success.
\newblock \emph{arXiv preprint arXiv:2607.12747}, 2026.

\bibitem[Yu et~al.(2025)Yu, Li, Xu, Fu, Hou, Lai, and Wang]{correct}
Y.~Yu, M.~Li, S.~Xu, J.~Fu, X.~Hou, F.~Lai, and B.~Wang.
\newblock {CORRECT}: {COndensed} {eRror} {RECognition} via knowledge transfer in multi-agent systems.
\newblock \emph{arXiv preprint arXiv:2509.24088}, 2025.

\bibitem[Zhang et~al.(2025{\natexlab{a}})Zhang, Wang, Chen, Zhou, Wang, and Yan]{agentracer}
G.~Zhang, J.~Wang, J.~Chen, W.~Zhou, K.~Wang, and S.~Yan.
\newblock {AgenTracer}: Who is inducing failure in the llm agentic systems?
\newblock \emph{arXiv preprint arXiv:2509.03312}, 2025{\natexlab{a}}.

\bibitem[Zhang et~al.(2025{\natexlab{b}})Zhang, Shi, Gu, You, Zhang, Gan, Yuan, and Huang]{graphtracer}
H.~Zhang, Y.~Shi, X.~Gu, H.~You, Z.~Zhang, L.~Gan, Y.~Yuan, and J.~Huang.
\newblock Graphtracer: Graph-guided failure tracing in llm agents for robust multi-turn deep search, 2025{\natexlab{b}}.
\newblock URL \url{https://arxiv.org/abs/2510.10581}.

\bibitem[Zhang et~al.(2025{\natexlab{c}})Zhang, Yin, Zhang, Liu, Han, Zhang, Li, Wang, Wang, Chen, and Wu]{whoandwhen}
S.~Zhang, M.~Yin, J.~Zhang, J.~Liu, Z.~Han, J.~Zhang, B.~Li, C.~Wang, H.~Wang, Y.~Chen, and Q.~Wu.
\newblock Which agent causes task failures and when? on automated failure attribution of llm multi-agent systems.
\newblock In \emph{International Conference on Machine Learning (ICML)}, 2025{\natexlab{c}}.

\bibitem[Zhu et~al.(2026{\natexlab{a}})Zhu, Hong, Wu, Chawla, Tang, Yin, Wolfe, Babinsky, and Liu]{raffles}
C.~Zhu, S.~Hong, J.~Wu, K.~Chawla, Y.~Tang, Y.~Yin, N.~Wolfe, E.~Babinsky, and D.~Liu.
\newblock {RAFFLES}: Reasoning-based attribution of faults for {LLM} systems.
\newblock In V.~Demberg, K.~Inui, and L.~Marquez, editors, \emph{Proceedings of the 19th Conference of the {E}uropean Chapter of the {A}ssociation for {C}omputational {L}inguistics (Volume 1: Long Papers)}, pages 7659--7688, Rabat, Morocco, Mar. 2026{\natexlab{a}}. Association for Computational Linguistics.
\newblock ISBN 979-8-89176-380-7.
\newblock \doi{10.18653/v1/2026.eacl-long.359}.
\newblock URL \url{https://aclanthology.org/2026.eacl-long.359/}.

\bibitem[Zhu et~al.(2025)Zhu, Liu, Li, Tian, Yang, Zhang, Han, Xie, Cui, Zhang, Ma, Yu, Ramesh, Wu, Liu, Lu, Zou, and You]{zhu2025agentdebug}
K.~Zhu, Z.~Liu, B.~Li, M.~Tian, Y.~Yang, J.~Zhang, P.~Han, Q.~Xie, F.~Cui, W.~Zhang, X.~Ma, X.~Yu, G.~Ramesh, J.~Wu, Z.~Liu, P.~Lu, J.~Zou, and J.~You.
\newblock Where {LLM} agents fail and how they can learn from failures.
\newblock \emph{arXiv preprint arXiv:2509.25370}, 2025.

\bibitem[Zhu et~al.(2026{\natexlab{b}})Zhu, Wu, Jin, Li, and Huang]{zhu2026stepfinder}
T.~Zhu, Y.~Wu, W.~Jin, Y.~Li, and G.~Huang.
\newblock {StepFinder}: A temporal semantic framework for failure attribution in multi-agent systems.
\newblock In \emph{Proceedings of the 32nd ACM SIGKDD Conference on Knowledge Discovery and Data Mining V. 2}, 2026{\natexlab{b}}.

\end{thebibliography}

\newpage
\appendix
\section{Additional experimental details}
\label{app:details}

\subsection{Datasets and statistics}
\label{app:datasets}

We evaluate on three failure-attribution benchmarks that together provide five reported subsets. Every trajectory is a failed run annotated with one decisive-error step and its responsible agent.

\noindent \textbf{\textsc{Who\&When}.}
We use the two subsets of \citet{whoandwhen}: \textsc{WW-AG} holds 126 trajectories generated by CaptainAgent, and \textsc{WW-HC} holds 58 trajectories of Magentic-One. 

\noindent \textbf{\textsc{Correct-Error}.}
The corpus of \citet{correct} spans seven question pools (ARC, GAIA, HotpotQA, MATH-500, MMLU-Pro, MuSiQue, and 2WikiMultihopQA), 2{,}226 trajectories in total. We report the column \textsc{CE} as the unweighted mean over the seven per-subset accuracies.

\noindent \textbf{\textsc{TraceElephant}.}
From \citet{traceelephant} we use the CaptainAgent subset \textsc{TE-Cap} (85
trajectories) and the Magentic-One subset \textsc{TE-Mag} (91 trajectories).
\textsc{TraceElephant} also releases, for every step, the context the agent saw when it
produced that step. We discard this context and keep only the step outputs, so its
trajectories take the same form as those of \textsc{Who\&When}.

\noindent \textbf{Trajectory statistics.}
Table~\ref{tab:datasets} reports the size and shape of every subset. Trajectory
length differs by an order of magnitude: a \textsc{WW-AG} trajectory has at most ten turns,
while a \textsc{WW-HC} trajectory averages 52 turns and reaches 130, which makes exact
localization hardest there.

\begin{table}[h]
\centering
\small
\setlength{\tabcolsep}{4.5pt}
\caption{Corpus statistics per subset: number of trajectories, turns per trajectory
(mean, median, maximum), distinct agents per trajectory, and the mean position of
the decisive error (0-indexed). The \textsc{CE} row is the unweighted mean over its seven
subsets.}

\label{tab:datasets}
\begin{tabular}{l l r rrr r r}
\toprule
Subset & Agent system & Traj. & \multicolumn{3}{c}{Turns} & Agents & Decisive step \\
\cmidrule(lr){4-6}
 & & & mean & med. & max & & mean \\
\midrule
\textsc{WW-AG}  & CaptainAgent & 126 & 8.7  & 10 & 10  & 3.6 & 3.0  \\
\textsc{WW-HC}  & Magentic-One & 58  & 51.6 & 32 & 130 & 5.7 & 15.2 \\
\textsc{TE-Cap} & CaptainAgent & 85  & 20.5 & 13 & 59  & 2.7 & 6.0  \\
\textsc{TE-Mag} & Magentic-One & 91  & 29.3 & 28 & 50  & 2.1 & 11.4 \\
\addlinespace[2pt]
\textsc{CE-ARC}      & CORRECT & 304 & 5.3  & 2  & 22 & 2.3 & 2.0 \\
\textsc{CE-GAIA}     & CORRECT & 50  & 10.9 & 10 & 22 & 2.9 & 7.7 \\
\textsc{CE-HotpotQA} & CORRECT & 578 & 12.8 & 10 & 22 & 3.0 & 5.2 \\
\textsc{CE-MATH-500} & CORRECT & 157 & 6.3  & 5  & 21 & 3.1 & 3.3 \\
\textsc{CE-MMLU-Pro} & CORRECT & 92  & 6.1  & 2  & 22 & 2.6 & 2.0 \\
\textsc{CE-MuSiQue}  & CORRECT & 312 & 14.7 & 21 & 22 & 3.0 & 6.1 \\
\textsc{CE-2WikiMQA} & CORRECT & 733 & 14.5 & 14 & 22 & 3.0 & 6.7 \\
\textsc{CE} (mean)   & CORRECT & 2{,}226 & 10.1 & 9.1 & 22 & 2.8 & 4.7 \\
\bottomrule
\end{tabular}
\end{table}

\subsection{Input format and context truncation}
\label{app:input}

\textbf{Input prompts.}
To score step $s_t$, the LLM $\proxy$ reads a single user turn of its chat template. The turn holds the task description followed by the steps $s_1,\ldots,s_t$. Each step is written as \texttt{[\{agent\}] - Step \{i\}: \{content\}}, and consecutive steps are separated by a blank line. No system prompt is used. The tokens of $s_t$ are used in both extracting its
representation $v_t$ and the attention mass $m_{i,t}$ (Appendix~\ref{app:implementation}).
In simplified form, the input to score $s_t$ reads:
\begin{methodbox}
\ttfamily\small
{[}\{agent$_1$\}{]} - Step 1: \{task description and first action\}\\[2pt]
{[}\{agent$_2$\}{]} - Step 2: \{content\}\\[2pt]
\ldots\\[2pt]
{[}\{agent$_t$\}{]} - Step $t$: \{content\}\hfill$\leftarrow$ scored step\\
\end{methodbox}

\noindent \textbf{Context truncation.}
The context budget is 8{,}192 tokens for every subset and backbone. The task
description always opens the context. Without the ground truth it is the question
alone; with the ground truth it is the question followed by the gold answer
(Appendix~\ref{app:gt}). When the task description, the preceding steps, and the
scored step together exceed the budget, we drop the oldest steps one at a time
until the input fits. The task description and the scored step are never dropped.

\subsection{Implementation details of \soap{}}
\label{app:implementation}

\textbf{Step representations.}
$v_t$ is the mean of the hidden states over the tokens of the scored step at one layer. We store the output of every full-attention block, the embedding layer, and the last block after the final norm; for architectures that interleave linear-attention and full-attention blocks (e.g., Qwen3.5), only the full-attention blocks are kept. The representation layer $l^\star$ is selected over these positions. Ablations on the pooling design are in Appendix~\ref{app:pooling}.

\noindent \textbf{Attention mass.}
Consider the input for step $s_t$ as the token sequence of $[s_1,\ldots,s_t]$, and let $\mathcal{T}_i$ denote the token positions of step $s_i$. Let $A^{(l,h)}$ be the post-softmax attention matrix of layer $l$ and head $h$: row $p$ holds the distribution of query token $p$ over the key tokens it may attend to. The mass that step $s_t$ routes to an earlier step $s_i$ is the probability that a typical query token of $s_t$ places on the tokens of $s_i$, averaged over the heads and over the layers of the selected band $L^\star$:
\begin{equation}
\small
    m^{(l,h)}_{i,t}
    =
    \frac{1}{|\mathcal{T}_t|}
    \sum_{p\in\mathcal{T}_t}\;\sum_{q\in\mathcal{T}_i}
    A^{(l,h)}_{p,q},
    \qquad
    m_{i,t}
    =
    \frac{1}{|L^\star|\,H}
    \sum_{l\in L^\star}\sum_{h=1}^{H}
    m^{(l,h)}_{i,t},
    \label{eq:attn-mass}
\end{equation}
where $H$ is the number of heads and  $L^\star$ is one of four equal bands of the full-attention blocks. Averaging over the query tokens treats the step as
a whole; summing over the key tokens keeps the total probability that lands in the
predecessor. Every $m_{i,t}$ for step $s_t$ comes from the forward pass that also
produces $v_t$:

\begin{methodbox}
\footnotesize
\textbf{Input:} steps $s_1,\ldots,s_t$; layer band $L^\star$. \quad
\textbf{Output:} $m_{i,t}$ for every $i<t$.\\[2pt]
    1.\ Run one forward pass of $\proxy$ on $[s_1,\ldots,s_t]$ and record the token span $\mathcal{T}_i$ of every step.\\
    2.\ \textbf{For} each layer $l\in L^\star$ and head $h$: take the attention rows of the query tokens in $\mathcal{T}_t$.\\
        \hspace*{1.4em}\textbf{For} each predecessor $i<t$: sum the probabilities over the key tokens in $\mathcal{T}_i$, then average over the query tokens; this is $m^{(l,h)}_{i,t}$.\\
    3.\ Average $m^{(l,h)}_{i,t}$ over the heads and the layers in $L^\star$ to obtain $m_{i,t}$.\\
    4.\ Normalize $m_{i,t}$ over the predecessor set $\mathcal{P}_t$ by Eq.~\ref{eq:dependency-weight}.
\end{methodbox}

\noindent \textbf{Spectral base score.}
The reference matrix $R$ stacks the representations of every step of the reference split, uncentered. We keep the 20 leading right-singular vectors and consider every contiguous band $[c_{\mathrm{begin}}, c_{\mathrm{end}})$ within this range.
The score is $\Sbase(s_t)=1/(q_{\mathcal{C}}(v_t)+\epsilon)$ with $\epsilon=10^{-12}$. Ranking is within a trajectory, and ties resolve to the earliest step.

\noindent \textbf{Construction of the predecessor set during rescoring.}
Section~\ref{sec:method:weights} normalizes the attention mass over an index set $\mathcal{P}_t$ of retained predecessors. We build $\mathcal{P}_t$ in three steps. First, context truncation (Appendix~\ref{app:input}) fixes which earlier steps the LLM reads when it encodes $s_t$; only those can receive attention from $s_t$. Second, among them we keep the $w$ predecessors with the largest mass $m_{i,t}$. The window $w\in\{1,2,3,4,5,\text{all}\}$ is selected per cell, and $w=\text{all}$ keeps every retained predecessor. Third, we remove any turn that is never a candidate step: the user's question turn of \textsc{WW-HC} and, with the ground truth, the pinned answer block. The weights $w_{i,t}$ of Eq.~\ref{eq:dependency-weight} are then normalized over the surviving set. Because $w_{i,t}=0$ outside $\mathcal{P}_t$, the sums in Eq.~\ref{eq:backward-evidence} run over the successors $t$ whose $\mathcal{P}_t$ contains $i$. A step that no successor retains keeps its base score, as does the last step. The propagation strength is selected from $\gamma\in\{0, 0.1, 0.2, 0.4, 0.6, 0.8, 1.0\}$, and $\gamma=0$ recovers the base score. Appendix~\ref{app:window} varies $w$.

\noindent \textbf{Hyperparameter selection.}
To keep the search space tractable, \soap{}'s hyperparameters are selected in two stages on the validation split. Stage~1 selects the representation layer $l^\star$ and the spectral band $\mathcal{C}$ by the validation step-level accuracy of the base score alone. Stage~2 freezes $(l^\star, \mathcal{C})$ and selects the attention band $L^\star$, propagation strength $\gamma$, and window $w$ by the validation step-level accuracy of the rescored prediction. Ties in both stages are broken by agent-level accuracy. Test labels are never consulted during selection. Baselines use the hyperparameters recommended in their respective papers.

\noindent \textbf{Alternative scoring functions.}
The seven alternatives of Table~\ref{tab:scorefn} share the selected layer and band width $|\mathcal{C}|$ of the spectral score; only the score itself changes. \emph{Perplexity} is the mean negative log-likelihood of the step's tokens under the LLM in context, read as higher~$=$~error. Four projection variants replace the spectral band with a different subspace of the same dimension: \emph{random subspace} uses a random orthonormal basis of dimension $|\mathcal{C}|$, \emph{top subspace} the leading band $[0, |\mathcal{C}|)$, \emph{tail subspace} the last $|\mathcal{C}|$ of the 20 computed components, and \emph{full spectrum} all 20. The $\ell_1$ and $\ell_2$ norms score $\|v_t\|_1$ and $\|v_t\|_2$. Every projection and norm score is read as lower~$=$~error, the same orientation as the spectral score. When the selected band starts at component~0, the top subspace \emph{is} the spectral band, which explains the ties in Table~\ref{tab:scorefn}.

\noindent \textbf{Alternatives to attention-guided rescoring.}
Every row of Table~\ref{tab:weights} starts from the same base score $\Sbase$ and
changes only how the correction is formed. The two content-agnostic rows keep the
form $\Stilde(s_i)=\Sbase(s_i)+\gamma B_i$ of Eq.~\ref{eq:recap}, with the selected
$\gamma$, and replace the dependency-weighted average of
Eq.~\ref{eq:backward-evidence} by a rule that ignores the attention:
\begin{equation}
\small
    B_i^{\mathrm{norm}}
    =
    \frac{1}{T-i}\sum_{t>i}\Sbase(s_t),
    \qquad
    B_i^{\mathrm{unnorm}}
    =
    \sum_{t>i}\Sbase(s_t).
    \label{eq:uniform-alt}
\end{equation}
The normalized form gives every successor the same weight; the unnormalized form
sums them, so an early step with many successors collects more. 

The three position
heuristics drop the propagation and act on the base score alone. \emph{Temporal
bias} subtracts a penalty that grows with the position of the step,
\begin{equation}
\small
    \Stilde^{\mathrm{base}}(s_t)
    =
    \Sbase(s_t)-\lambda\,\frac{t}{T},
    \qquad
    \Stilde^{\mathrm{std}}(s_t)
    =
    z\bigl(\Sbase(s_t)\bigr)-\lambda\,\frac{t}{T},
    \label{eq:temporal-alt}
\end{equation}
where $z(\cdot)$ standardizes the base scores within the trajectory to zero mean and
unit variance, and $\lambda$ is selected per column from $\{0.1, 0.2, \ldots, 1.0\}$
by the same rule as every other hyperparameter. \emph{Earliest of top-$5$} predicts
the smallest index among the five steps with the largest base score. Each heuristic
pushes blame toward earlier steps without asking which earlier step the later ones
depend on, which is the comparison Table~\ref{tab:weights} makes.

\subsection{Baselines}
\label{app:baselines}

\textbf{Prompt-based methods.}
All-at-Once, Step-by-Step, and Binary Search run the code released with \textsc{Who\&When}~\citep{whoandwhen}. CORRECT~\citep{correct} and CHIEF~\citep{chief} are reproduced from their released code with the prompts kept unchanged. ErrorProbe~\citep{errorprobe} runs the pipeline of its paper: a failure-mode tagger and a dependency graph over chunks of the trajectory, then a Strategist, an Investigator, and an Arbiter that names the decisive step. For every method we adopt the decoding settings its paper recommends. The with-GT and without-GT runs differ in one line of the prompt, which states the gold answer. We did not run the GPT-5 judge on the \textsc{Correct-Error} corpus, whose 2{,}226 trajectories and multi-call protocols made it the most expensive cell of the sweep; Appendix~\ref{app:prompting-gpt5} therefore omits it.

\noindent \textbf{OAT.}
OAT models the latent dynamics of \emph{successful} trajectories and scores each step of a failed trajectory by how far it departs from them. A neural controlled differential equation predicts each step's latent state from the trajectory so far; the anomaly score is the squared distance between the predicted and observed states, and the predicted decisive step is the argmax. Training is one-class on the successful tool-use trajectories released with the paper (\textsc{MCP-Atlas}), following the paper's own out-of-distribution protocol; one trained model serves every subset. In the with-GT setting the gold answer is appended to the task description with the same phrasing the prompt-based baselines use; the training corpus carries no answers, so the model is shared between settings.

\noindent \textbf{StepFinder.}
StepFinder is a supervised step classifier. Each step's text and agent name are embedded separately, and the sequence passes through a BiLSTM, an agent-aware attention layer, and a scoring head with a position prior, trained with a per-trajectory cross-entropy loss plus a self-supervised next-step objective. Training uses the step-labeled failure corpora released with the paper. To place StepFinder on the same backbones as \soap{}, we swap its original encoder for the backbone's last-layer hidden states. In the with-GT setting the gold answer is appended to the first step, since StepFinder encodes no separate task description; the models are shared between settings.

\noindent \textbf{Evaluation.}
OAT and StepFinder score steps but name no agent; their responsible-agent prediction is the agent of the predicted step. The prompt-based judges name an agent directly, as the \textsc{Who\&When} prompts request, and that name is scored. We train five models per backbone; a reported cell averages over the subset's three test splits and, within each split, the five training seeds. All methods follow the same scoring rules as Table~\ref{tab:main}. 

\subsection{With-ground-truth setting}
\label{app:gt}

\textbf{How \soap{} receives the answer.}
In the with-GT setting, the LLM's context opens with a pinned block, \textit{``The problem is: \{question\}. The answer for the problem is: \{answer\}''}, phrased as in the prompting baselines so the LLM reads the answer in the same register. The block precedes every turn and survives context truncation. OAT appends the answer to the task description and StepFinder to the first step, as described in Appendix~\ref{app:baselines}.

\noindent \textbf{Full results.}
Table~\ref{tab:gt-full} extends Table~\ref{tab:main-gt} to all five subsets, the GPT-4o judge, and DeepSeek-8B. \soap{} is best on every column of both open-backbone blocks, tied only by Step-by-Step on \textsc{TE-Mag} with Qwen3.5-9B (21.01). The answer helps the judges more than it helps \soap{}: Step-by-Step gains 21 points on \textsc{CE} with Qwen3.5-9B and 15 with GPT-4o, while \soap{} moves by under 5 points everywhere except DeepSeek-8B \textsc{TE-Mag} (+6.5). The closest a judge gets is on \textsc{CE} with the answer, where Binary Search and ErrorProbe with Qwen3.5-9B reach 59.89 and 59.57 against \soap{}'s 60.59.

\begin{table}[h]
\centering
\footnotesize
\setlength{\tabcolsep}{5pt}
\caption{Step-level accuracy (\%) in the with-ground-truth setting on all five
subsets. The prompt-based rows use the judge named in the block; the training-based
rows and \soap{} use the backbone named in the block. The
best result in each column within a block is highlighted in \textbf{bold}.}
\label{tab:gt-full}
\begin{tabular}{l ccccc}
\toprule
Method & \textsc{WW-AG} & \textsc{WW-HC} & \textsc{CE} & \textsc{TE-Cap} & \textsc{TE-Mag} \\
\midrule
\rowcolor{Gray}\multicolumn{6}{c}{\textit{\textbf{GPT-4o}}} \\
All-at-Once   & 15.87 & 3.45  & 24.50 & 9.30  & 6.52 \\
Step-by-Step  & 22.75 & 18.39 & 59.29 & 18.60 & 19.57 \\
Binary Search & 28.57 & 4.60  & 18.44 & 9.30  & 6.52 \\
CORRECT       & 15.87 & 4.60  & 37.09 & 10.85 & 10.14 \\
CHIEF         & 24.34 & 11.49 & 42.10 & 14.73 & 13.77 \\

ErrorProbe    & \textcolor{gray}{\best{34.92}} & \textcolor{gray}{\best{28.74}} & \textcolor{gray}{\best{62.26}} & \textcolor{gray}{\best{30.23}} & \textcolor{gray}{\best{26.81}} \\
\midrule
\rowcolor{Gray}\multicolumn{6}{c}{\textit{\textbf{Qwen3.5-9B}}} \\

All-at-Once   & 21.16 & 11.49 & 33.32 & 6.98  & 1.45 \\
Step-by-Step  & 17.46 & 18.39 & 52.04 & 24.81 & \best{21.01} \\
Binary Search & 2.12  & 13.79 & 59.89 & 0.00  & 10.14 \\
CORRECT       & 32.28 & 4.60  & 36.43 & 10.08 & 8.70 \\
CHIEF         & 30.69 & 4.60  & 30.78 & 6.98  & 10.87 \\
ErrorProbe    & 40.74 & 19.54 & 59.57 & 24.81 & 20.29 \\
StepFinder    & 18.52 & 12.87 & 28.52 & 13.64 & 8.99 \\
OAT           & 22.12 & 8.05  & 56.21 & 17.98 & 20.00 \\
\rowcolor{gray!10}\soap{} (w/o rescoring) & 41.27 & 33.33 & 60.30 & 32.56 & 15.94 \\
\rowcolor{gray!10}\soap{}       & \best{43.39} & \best{34.48} & \best{60.59} & \best{36.43} & \best{21.01} \\
\midrule
\rowcolor{Gray}\multicolumn{6}{c}{\textit{\textbf{DeepSeek-R1-Distill-Llama-8B}}} \\
All-at-Once   & 21.16 & 5.75  & 24.16 & 8.53  & 4.35 \\
Step-by-Step  & 15.34 & 3.45  & 16.51 & 10.85 & 5.07 \\
Binary Search & 6.35  & 12.64 & 40.01 & 4.65  & 7.25 \\
CORRECT       & 19.58 & 11.49 & 36.19 & 14.73 & 1.45 \\
CHIEF         & 14.29 & 11.49 & 10.95 & 4.65  & 0.00 \\
ErrorProbe    & 18.52 & 14.94 & 40.39 & 21.71 & 11.59 \\
StepFinder    & 18.94 & 9.89  & 24.99 & 15.04 & 6.23 \\
OAT           & 20.85 & 17.01 & 52.39 & 18.14 & 13.77 \\
\rowcolor{gray!10} \soap{} (w/o rescoring) & 40.21 & 29.89 & 64.81 & 41.09 & 35.51 \\
\rowcolor{gray!10}\soap{}       & \best{44.97} & \best{29.89} & \best{65.19} & \best{42.64} & \best{36.96} \\
\bottomrule
\end{tabular}
\end{table}

\subsection{Compute resources and cost}
\label{app:compute}

We compare execution time on the frozen Qwen3.5-9B backbone. All experiments use Python 3.10, PyTorch 2.13, and Transformers 5.15 on one NVIDIA H200 (141\,GB), processing one trajectory at a time.

\noindent \textbf{Fitting and inference time.}
Table~\ref{tab:cost-inference} reports fitting time, forward-pass time, and CPU scoring time per trajectory. \soap{} fits its SVD on the unlabeled reference set in 0.08\,s on \textsc{WW-AG} and 0.29\,s on \textsc{WW-HC}, two to three orders of magnitude faster than OAT and StepFinder (51--115\,s per training seed). One forward pass costs 0.38\,s on \textsc{WW-AG} and 0.76\,s on \textsc{WW-HC}, on par with OAT (0.42 and 2.13\,s) and StepFinder (0.23 and 0.19\,s). Scoring the stored vectors then takes 0.7 and 5.5\,ms per trajectory, the fastest of the three.

\begin{table}[htbp]
\centering
\footnotesize
\setlength{\tabcolsep}{12pt}
\renewcommand{\arraystretch}{1.12}
\caption{Fitting and inference time on \textsc{Who\&When} with Qwen3.5-9B. Fitting is per training seed (mean of five) for OAT and StepFinder, one CPU SVD fit for \soap{}. Extraction is one forward pass of the backbone.}
\label{tab:cost-inference}
\resizebox{\ifdim\width>\textwidth \textwidth\else\width\fi}{!}{%
\begin{tabular}{l rr rr rr}
\toprule
& \multicolumn{2}{c}{\shortstack{Fitting\\(s/fit)}}
& \multicolumn{2}{c}{\shortstack{Embedding extraction\\(s/forward pass)}}
& \multicolumn{2}{c}{\shortstack{CPU scoring\\(ms/trajectory)}} \\
\cmidrule(lr){2-3}\cmidrule(lr){4-5}\cmidrule(lr){6-7}
Method
& \textsc{WW-AG} & \textsc{WW-HC}
& \textsc{WW-AG} & \textsc{WW-HC} 
& \textsc{WW-AG} & \textsc{WW-HC} \\
\midrule
OAT
& 51.10 & 51.10
& 0.42 & 2.13 
& 2.4 & 10.4 \\
StepFinder
& 51.79 & 115.43
& {0.23} & {0.19} 
& 1.8 & 5.0 \\
\rowcolor{gray!10}
\soap{}
& \textbf{0.08} & \textbf{0.29}
& 0.38 & 0.76 
& \textbf{0.7} & \textbf{5.5} \\
\bottomrule
\end{tabular}}
\end{table}

\noindent \textbf{Comparison with prompt-based methods.}
Table~\ref{tab:cost-judges} compares all methods on the same 40 \textsc{WW-AG} and 17 \textsc{WW-HC} trajectories with Qwen3.5-9B, without the ground-truth answer. Each judge is timed over all of its requests per trajectory, prompt processing and decoding included; \soap{} is timed over one forward pass plus CPU scoring. \soap{} takes 0.38\,s on \textsc{WW-AG} and 0.76\,s on \textsc{WW-HC}, faster than every judge except Binary Search (0.2 and 0.4\,s). The multi-call judges are the slowest: CHIEF takes 24\,s and ErrorProbe 6\,s per trajectory.

\begin{table}[htbp]
\centering
\footnotesize
\setlength{\tabcolsep}{8pt}
\caption{Time per trajectory (s) on the shared \textsc{Who\&When} subset.}
\label{tab:cost-judges}
\begin{tabular}{l rr}
\toprule
Method & AG & HC \\
\midrule
All-at-Once   & 2.0  & 2.1 \\
Step-by-Step  & 2.0  & 3.6 \\
Binary Search & 0.2  & 0.4 \\
CORRECT       & 1.9  & 1.6 \\
CHIEF         & 23.9 & 23.5 \\
ErrorProbe    & 5.9  & 6.5 \\
\midrule
\rowcolor{gray!10}
\soap{}       & 0.38 & 0.76 \\
\bottomrule
\end{tabular}
\end{table}

\subsection{Synthetic reference trajectories}
\label{app:synth}

The synthetic references of Figure~\ref{fig:transfer}(d) come from re-running the agent systems behind the two \textsc{Who\&When} subsets, CaptainAgent for \textsc{WW-AG} and Magentic-One for \textsc{WW-HC}, once with Qwen3.5-9B and once with GPT-4o as the backbone of their agents. The systems run on the same question pools the benchmarks draw from (GAIA and AssistantBench), and we keep the trajectories whose question belongs to the target subset. The runs are unsupervised: the corpus mixes successes and failures, carries no step-level annotation, and serves only to construct $R$. The existing benchmarks and our synthetic trajectories share a question pool but never a trajectory.

The evaluation follows Table~\ref{tab:main}; only the reference changes. Our GPT-4o corpus differs from the \textsc{Who\&When} trajectories despite the shared model name for two reasons: \textsc{Who\&When} was generated with the \texttt{2024-08-01-preview} version of GPT-4o, whereas ours uses \texttt{2024-08-06}, and we rebuilt the generation pipeline from the paper's description because the benchmark does not release its generation code. Appendix~\ref{app:synth-results} reports the full results for both backbones.

\section{Additional ablation studies}
\label{app:deepseek-ablations}

The main text ablates \soap{} on \textsc{WW-AG} and \textsc{WW-HC} with backbone Qwen3.5-9B (Section~\ref{sec:exp:analysis}). This appendix extends the ablations to the remaining cells, \textsc{TE-Cap} and \textsc{TE-Mag} on Qwen3.5-9B and all four subsets on DeepSeek-8B, under the same protocol: start from the configuration selected for Table~\ref{tab:main}, vary one axis, hold everything else fixed.

\subsection{Context window of the dependency weights}
\label{app:window}

Table~\ref{tab:attnsel} varies the window $w$, which controls the number of predecessors each step distributes its blame over, with everything else at the selected configuration. The preference is dataset-dependent. On \textsc{Who\&When} both backbones favor the sharpest window: accuracy peaks at $w{=}1$--$2$ and declines toward $w{=}\text{all}$ (47.62 vs.\ 40.21 on Qwen3.5-9B \textsc{WW-AG}), because restricting each step to its strongest dependencies keeps the correction targeted rather than diluting it toward the uniform weights of Table~\ref{tab:weights}. On \textsc{TE-Cap} the pattern reverses: wider windows win on both backbones, up to $w{=}\text{all}$ on DeepSeek-8B. A poorly chosen $w$ can fall below the base score (Qwen3.5-9B \textsc{TE-Mag} at $w{=}1$: 15.94 vs.\ 21.01), confirming that $w$ is worth selecting rather than fixing.

\begin{table}[h]
\centering
\small
\setlength{\tabcolsep}{5pt}
\caption{Context window $w$ of the dependency weights. Step-level accuracy (\%) when
each step keeps its $w$ strongest predecessors, everything else at the selected
configuration; the selected $w$ is highlighted in \textbf{bold}. DeepSeek-8B's
selected $\gamma$ on \textsc{WW-HC} is 0, so that row is constant by construction.}
\label{tab:attnsel}
\begin{tabular}{ll cccccc}
\toprule
Backbone & Subset & $w=1$ & $2$ & $3$ & $4$ & $5$ & all \\
\midrule
\multirow{4}{*}{Qwen3.5-9B}
 & \textsc{WW-AG}  & \best{47.62} & 41.80 & 40.74 & 40.21 & 40.21 & 40.21 \\
 & \textsc{WW-HC}  & 33.33 & \best{34.48} & 32.18 & 32.18 & 32.18 & 32.18 \\
 & \textsc{TE-Cap} & 28.68 & 29.46 & 34.88 & 34.88 & \best{35.66} & 34.88 \\
 & \textsc{TE-Mag} & 15.94 & 20.29 & 18.84 & \best{23.19} & 18.84 & 18.12 \\
\midrule
\multirow{4}{*}{DeepSeek-8B}
 & \textsc{WW-AG}  & \best{45.50} & 38.62 & 33.33 & 34.39 & 37.57 & 37.57 \\
 & \textsc{WW-HC}  & 28.74 & 28.74 & 28.74 & 28.74 & 28.74 & \best{28.74} \\
 & \textsc{TE-Cap} & 35.66 & 38.76 & 41.09 & 41.09 & 41.86 & \best{42.64} \\
 & \textsc{TE-Mag} & \best{30.43} & 30.43 & 30.43 & 29.71 & 28.99 & 28.99 \\
\bottomrule
\end{tabular}
\end{table}

\subsection{Sensitivity to the hyperparameters on all cells}
\label{app:sensitivity}

\begin{figure}[h]
\centering
\includegraphics[width=\textwidth]{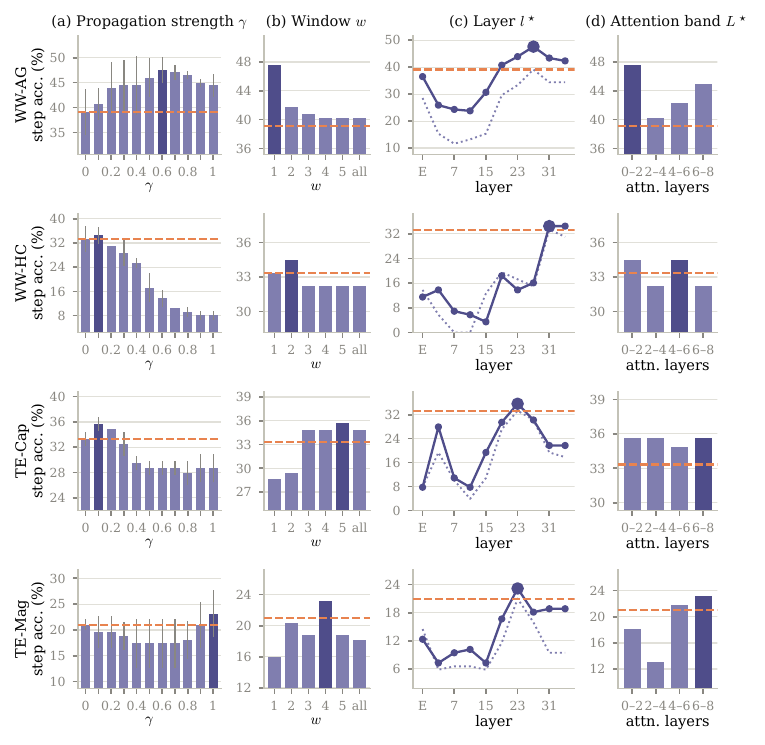}
\caption{Sensitivity of \soap{} on Qwen3.5-9B, all four subsets. As
Figure~\ref{fig:datasize}, with the window $w$ added: (a) propagation strength
$\gamma$, (b) window $w$, (c) representation layer $l^\star$ (solid: after rescoring;
dotted: base score at that layer; large marker: selected layer), (d) attention band
$L^\star$. The dashed orange line denotes the base score of the selected
configuration.}
\label{fig:sensitivity-qwen}
\end{figure}

\begin{figure}[h]
\centering
\includegraphics[width=\textwidth]{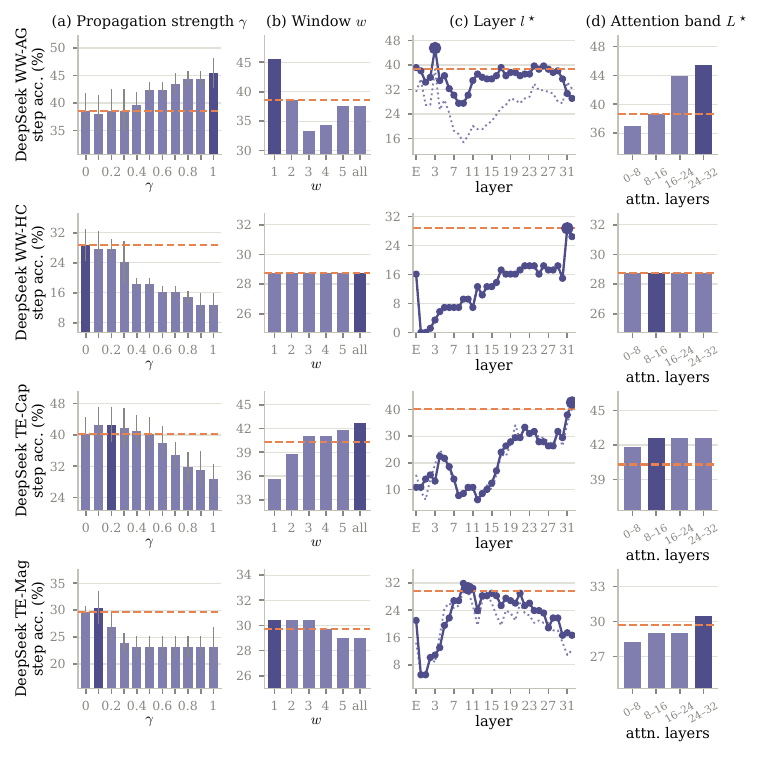}
\caption{Sensitivity of \soap{} on DeepSeek-8B, all four subsets;
panels as in Figure~\ref{fig:sensitivity-qwen}. DeepSeek-8B sweeps every block in
panel (c). On \textsc{WW-HC} the selected $\gamma$ is 0, so the $w$ and attention-band panels
are flat by construction.}
\label{fig:sensitivity-deepseek}
\end{figure}

Figures~\ref{fig:sensitivity-qwen} and \ref{fig:sensitivity-deepseek} repeat
Figure~\ref{fig:datasize} on all eight (backbone, subset) cells and add the window
$w$ as a panel. Two regularities hold across backbones. First, $\gamma$ has the same
two regimes: \textsc{WW-AG} climbs toward large $\gamma$ on DeepSeek-8B as on Qwen3.5-9B
(38.62 at $\gamma=0$ to 45.50 at $\gamma=1.0$), while every other cell peaks at
$\gamma \in \{0.1, 0.2\}$ and decays beyond, so the correction is a nudge where the
base score is already strong and carries most of the signal where it is not. Second,
the selected layer is the per-layer optimum of post-rescoring accuracy in seven of
the eight cells (the exception is DeepSeek-8B \textsc{TE-Mag}, where block 9 edges the
selected block 10 by 1.4 points); \textsc{WW-HC} concentrates in the last block on both
backbones, and DeepSeek-8B consistently prefers the late attention band (24--32).
These results suggest that the selected configurations sit on broad optima rather
than on isolated peaks.

\subsection{Pooling of the step representation}
\label{app:pooling}

\soap{} averages the hidden states over a step's tokens. We compare this with using the
last-token representation. As shown in Table~\ref{tab:pooling}, mean
pooling is clearly better: on \textsc{WW-AG} the base score drops from 39.15 to 29.10 on
Qwen3.5-9B and from 38.62 to 29.10 on DeepSeek-8B, and on \textsc{TE-Cap} from 40.31 to 24.81
on DeepSeek-8B; only \textsc{TE-Mag} is close. Rescoring still helps on last-token
representations (e.g., 29.10 to 30.69 on DeepSeek-8B \textsc{WW-AG}), but from a much lower
base. These results indicate that the spectral signal lives in the step's tokens as a
whole rather than in the state at its final token, which mostly reflects what comes
next.

\begin{table}[h]
\centering
\footnotesize
\setlength{\tabcolsep}{3.5pt}
\caption{Mean vs.\ last-token pooling of the step representation. We compare the step-level accuracy
(\%); the best result in each
column is highlighted in \textbf{bold}.}
\label{tab:pooling}
\begin{tabular}{l l cccc cccc}
\toprule
 & & \multicolumn{4}{c}{Qwen3.5-9B} & \multicolumn{4}{c}{DeepSeek-8B} \\
\cmidrule(lr){3-6}\cmidrule(lr){7-10}
Pooling & Method & WW-AG & WW-HC & TE-Cap & TE-Mag & WW-AG & WW-HC & TE-Cap & TE-Mag \\
\midrule
Last token & Base  & 29.10 & 22.99 & 24.03 & 21.01 & 29.10 & 22.99 & 24.81 & 24.64 \\
           & \soap & 29.63 & 24.14 & 25.58 & 23.19 & 30.69 & 22.99 & 24.81 & 26.09 \\
\rowcolor{Gray}
Mean (Ours) & Base & 39.15 & 33.33 & 33.33 & 21.01 & 38.62 & 28.74 & 40.31 & 29.71 \\
\rowcolor{Gray}
            & \soap & \best{47.62} & \best{34.48} & \best{35.66} & \best{23.19} & \best{45.50} & \best{28.74} & \best{42.64} & \best{30.43} \\
\bottomrule
\end{tabular}
\end{table}

\subsection{Quantity of unlabeled reference data}
\label{app:datasize}

Table~\ref{tab:datasize-app} completes Figure~\ref{fig:datasize}(d) with the
remaining cells and DeepSeek-8B. With a third of the reference corpus (6--12
trajectories) every cell is within about four points of its full-data accuracy, and
most within one or two. Because the configuration is re-selected at each fraction on
the validation split, a small fit can occasionally score above the full one (DeepSeek-8B
\textsc{WW-HC} and \textsc{TE-Mag} at 10\%); this is selection noise on tiny fits, not a real gain.
These results confirm that a handful of unlabeled trajectories suffices to construct
the reference.

\begin{table}[h]
\centering
\small
\setlength{\tabcolsep}{6pt}
\caption{Quantity of unlabeled reference data, all cells. Step-level accuracy (\%) of
\soap{} when the reference is fit on $\{10, 20, 30\}\%$ of the corpus, validation and
test fixed, full configuration re-selected per fraction. The best result in each column
within a block is highlighted in \textbf{bold}.}
\label{tab:datasize-app}
\begin{tabular}{l cccc}
\toprule
Reference data & \textsc{WW-AG} & \textsc{WW-HC} & \textsc{TE-Cap} & \textsc{TE-Mag} \\
\midrule
\rowcolor{Gray}\multicolumn{5}{c}{\textit{\textbf{Qwen3.5-9B}}} \\
\quad 10\% & 43.39 & 33.33 & 34.88 & \best{23.19} \\
\quad 20\% & 44.44 & 33.33 & 34.88 & \best{23.19} \\
\quad 30\% & \best{47.62} & \best{34.48} & \best{35.66} & \best{23.19} \\
\midrule
\rowcolor{Gray}\multicolumn{5}{c}{\textit{\textbf{DeepSeek-R1-Distill-Llama-8B}}} \\
\quad 10\% & 37.04 & \best{29.89} & 41.86 & \best{31.88} \\
\quad 20\% & \best{46.03} & 28.74 & \best{42.64} & 30.43 \\
\quad 30\% & 45.50 & 28.74 & \best{42.64} & 30.43 \\
\bottomrule
\end{tabular}
\end{table}

\subsection{Length and norm baselines}
\label{app:length}

A natural concern is that the base score only tracks surface properties of a step, such as its length or the norm of its representation. To test this, we compare \soap{} with two trivial baselines, which predict the longest or the shortest step, and with a unit-norm control that scales every step representation to $\lVert v_t \rVert_2 = 1$ in both the reference and test sets. The control either keeps the spectral band selected on the original representations (\emph{anchor band}) or re-selects it on the normalized ones (\emph{re-selected}), since the best band may shift once the norm is removed. As shown in Table~\ref{tab:length}, length alone does not locate the decisive step: the longest step scores only 8--20\%, 9--32 points below the base score, and the shortest step falls below random on every cell. Removing the norm lowers accuracy but leaves a clear signal: with the re-selected band, the unit-norm score reaches 22--35\%, above every length and random row, and trails the base score by 3--13 points on seven of the eight cells, even exceeding it on Qwen3.5-9B \textsc{TE-Mag} (23.19 vs.\ 21.01). \textsc{WW-AG} loses under 5 points on both backbones, because its selected band skips the top singular vector, which carries most of the norm variation. \soap{} achieves the best accuracy on every cell (e.g., 47.62 vs.\ 35.45 for the best unit-norm score on Qwen3.5-9B \textsc{WW-AG}), tied only by the base score on DeepSeek-8B \textsc{WW-HC}, where the selected $\gamma$ is 0. These results confirm that the base score captures a directional signal beyond step length and representation norm, and that rescoring builds on this signal.

\begin{table}[h]
\centering
\small
\setlength{\tabcolsep}{5pt}
\caption{Trivial baselines and the unit-norm control. Step-level accuracy (\%) on the
test split; the best result in each column is \textbf{bold}.}
\label{tab:length}
\begin{tabular}{l cccc}
\toprule
 & \textsc{WW-AG} & \textsc{WW-HC} & \textsc{TE-Cap} & \textsc{TE-Mag} \\
\midrule
\rowcolor{Gray}\multicolumn{5}{c}{\textit{\textbf{Qwen3.5-9B}}} \\
Longest step         & 19.05 & 19.54 & 10.08 & 11.59 \\
Shortest step        &  0.53 &  0.00 &  4.65 &  2.90 \\
Random               & 12.10 &  5.08 &  8.58 &  5.16 \\
Unit-norm, anchor band      & 35.45 & 16.09 & 24.81 & 16.67 \\
Unit-norm, re-selected      & 35.45 & 22.99 & 30.23 & 23.19 \\
Base score           & 39.15 & 33.33 & 33.33 & 21.01 \\
\soap{}              & \best{47.62} & \best{34.48} & \best{35.66} & \best{23.19} \\
\midrule
\rowcolor{Gray}\multicolumn{5}{c}{\textit{\textbf{DeepSeek-R1-Distill-Llama-8B}}} \\
Longest step         & 15.87 & 19.54 &  8.53 & 12.32 \\
Shortest step        &  0.53 &  0.00 &  4.65 &  1.45 \\
Random               & 12.10 &  5.08 &  8.58 &  5.16 \\
Unit-norm, anchor band      & 29.10 &  6.90 & 25.58 & 10.87 \\
Unit-norm, re-selected      & 33.86 & 22.99 & 27.13 & 22.46 \\
Base score           & 38.62 & 28.74 & 40.31 & 29.71 \\
\soap{}              & \best{45.50} & \best{28.74} & \best{42.64} & \best{30.43} \\
\bottomrule
\end{tabular}
\end{table}

\subsection{Effect of the reference corpus}
\label{app:reference-controls}

\soap{} constructs its spectral reference from the step-unlabeled failure trajectories of the benchmark's reference split. We further examine how the choice of this corpus affects attribution by replacing it with five alternatives. \emph{Random basis} replaces the reference directions with a random orthonormal basis of the same dimension as the selected band, which encodes no textual structure. \emph{\textsc{MCP-Atlas} successes} and \emph{\textsc{MCP-Atlas} failures} each draw 50 runs from \textsc{MCP-Atlas}~\citep{bandi2026mcpatlas}, a corpus of 198 single-agent tool-use runs by Qwen3.5-27B, and therefore serve as out-of-distribution references. \emph{Steps before $t^\star$} fits $R$ only on the reference steps that precede the annotated decisive step, the closest approximation to a success-only reference that the benchmarks allow. \emph{Random subset} draws the same number of steps uniformly from the full reference split, which separates the effect of using fewer steps from that of using earlier, cleaner steps. The last two references read step labels and serve only as diagnostics. For every reference, we re-select the configuration on the validation split and report the step-level accuracy on the test split.

\begin{table}[h]
\centering
\small
\setlength{\tabcolsep}{5pt}
\caption{Effect of the reference corpus on \textsc{Who\&When}. Step-level accuracy (\%) on the test split. $^\dagger$Diagnostic references that read step labels on the reference split. The best result in each column within a block is highlighted in \textbf{bold}.}
\label{tab:ref-controls}
\begin{tabular}{l cc}
\toprule
Reference & \textsc{WW-AG} & \textsc{WW-HC} \\
\midrule
\rowcolor{Gray}\multicolumn{3}{c}{\textit{\textbf{Qwen3.5-9B}}} \\
Random basis                   & 33.33        & 25.29 \\
\textsc{MCP-Atlas} successes            & 38.10        & 28.74 \\
\textsc{MCP-Atlas} failures             & 42.86        & 28.74 \\
Steps before $t^\star$$^\dagger$ & 36.51        & 33.33 \\
Random subset$^\dagger$        & 39.68        & 33.33 \\
Benchmark reference split      & \best{47.62} & \best{34.48} \\
\midrule
\rowcolor{Gray}\multicolumn{3}{c}{\textit{\textbf{DeepSeek-R1-Distill-Llama-8B}}} \\
Random basis                   & 36.51        & 27.59 \\
\textsc{MCP-Atlas} successes            & 40.21        & 29.89 \\
\textsc{MCP-Atlas} failures             & 44.44        & \best{32.18} \\
Steps before $t^\star$$^\dagger$ & 44.97        & 31.03 \\
Random subset$^\dagger$        & 39.15        & 27.59 \\
Benchmark reference split      & \best{45.50} & 28.74 \\
\bottomrule
\end{tabular}
\end{table}

As shown in Table~\ref{tab:ref-controls}, the benchmark reference performs best on \textsc{WW-AG} with both backbones, outperforming every alternative by up to 14.29 points with Qwen3.5-9B and 8.99 points with DeepSeek-8B. Moreover, failed trajectories provide a more informative reference than successful ones: \textsc{MCP-Atlas} failures outperform \textsc{MCP-Atlas} successes on \textsc{WW-AG} with both backbones (e.g., 42.86 vs.\ 38.10 with Qwen3.5-9B) and match or exceed them on \textsc{WW-HC}. Removing the decisive step and its downstream steps also hurts with Qwen3.5-9B, lowering \textsc{WW-AG} accuracy from 47.62 to 36.51, below even a random subset of the same size (39.68); with DeepSeek-8B, the steps before $t^\star$ nearly match the full reference (44.97 vs.\ 45.50). The random basis performs worst on \textsc{WW-AG}, 9--14 points below the benchmark reference. On \textsc{WW-HC} the differences narrow, and with DeepSeek-8B two alternatives even exceed the benchmark reference (32.18 for \textsc{MCP-Atlas} failures and 31.03 for the steps before $t^\star$, vs.\ 28.74). These results indicate that failed trajectories carry more informative signal for decisive-error localization.

\subsection{Comparison with a supervised probe}
\label{app:probe}

\soap{} uses the step-level labels of the 20\% validation split only to select its configuration. We further examine whether a supervised classifier trained directly on these labels can localize the decisive error more accurately. The probe is a logistic-regression classifier on the 4096-dimensional mean-pooled hidden state, with one example per step and balanced class weights. It is trained on the validation split alone, and its layer and regularization strength are selected by leave-one-trajectory-out cross-validation within that split. Like \soap{}, the probe predicts the step with the largest score in each trajectory, and we report the step-level accuracy on the test split.

\begin{table}[h]
\centering
\small
\setlength{\tabcolsep}{5pt}
\caption{Comparison with a supervised probe trained on validation labels. Step-level accuracy (\%) on the test split. The best result in each column within a block is highlighted in \textbf{bold}.}
\label{tab:probe}
\begin{tabular}{l cccc}
\toprule
 & \textsc{WW-AG} & \textsc{WW-HC} & \textsc{TE-Cap} & \textsc{TE-Mag} \\
\midrule
\rowcolor{Gray}\multicolumn{5}{c}{\textit{\textbf{Qwen3.5-9B}}} \\
Probe         & 43.92        & 20.69        & 18.60        & 17.39 \\
\soap{}       & \best{47.62} & \best{34.48} & \best{35.66} & \best{23.19} \\
\midrule
\rowcolor{Gray}\multicolumn{5}{c}{\textit{\textbf{DeepSeek-R1-Distill-Llama-8B}}} \\
Probe         & 35.45        & 19.54        & 20.93        & 20.29 \\
\soap{}       & \best{45.50} & \best{28.74} & \best{42.64} & \best{30.43} \\
\bottomrule
\end{tabular}
\end{table}

As shown in Table~\ref{tab:probe}, \soap{} outperforms the probe on every subset with both backbones, by 3.70--17.06 points with Qwen3.5-9B and 9.20--21.71 points with DeepSeek-8B. The probe comes closest on \textsc{WW-AG} with Qwen3.5-9B (43.92 vs.\ 47.62), the subset with the largest validation split. The largest gap arises on \textsc{TE-Cap} with both backbones (e.g., 20.93 vs.\ 42.64 with DeepSeek-8B), suggesting that a probe fitted on a few labeled trajectories overfits them. These results indicate that, when only a few labeled trajectories are available, using them to select the configuration of \soap{} is more effective than training a classifier on them.

\section{Additional analysis}
\label{app:analysis}

\subsection{Agent-level accuracy and variance across seeds}
\label{app:metrics}

\textbf{Agent-level accuracy.}
Table~\ref{tab:agent} reports the responsible-agent accuracy of every row of Table~\ref{tab:main} on \textsc{Who\&When}, i.e., how often the agent of the predicted step is the annotated agent. The metric is far more forgiving than step-level accuracy because a trajectory involves 3.6 agents on \textsc{WW-AG} and 5.7 on \textsc{WW-HC} on average (Table~\ref{tab:datasets}), so a method that names the right agent at a wrong step still scores. \soap{} is best on both subsets with both open-weight backbones (60.32 on \textsc{WW-AG} and 67.82 on \textsc{WW-HC} with Qwen3.5-9B; 61.90 and 65.52 with DeepSeek-8B). The prompt-based judges close much of their step-level gap here: All-at-Once with GPT-4o reaches 59.77 on \textsc{WW-HC} while locating the step in 6.90\% of cases, which says that a judge often blames the right agent for the wrong reason. These results indicate that agent-level accuracy rewards the coarse verdict a judge produces naturally, whereas step-level accuracy measures localization.

\begin{table}[h]
\centering
\footnotesize
\setlength{\tabcolsep}{5pt}
\caption{Agent-level accuracy (\%) in the without-ground-truth setting on
\textsc{Who\&When}, for every row of Table~\ref{tab:main}. Mean over the seed triple. The
best result in each column within a block is highlighted in \textbf{bold}.  }
\label{tab:agent}
\begin{tabular}{l cc}
\toprule
Method & \textsc{WW-AG} & \textsc{WW-HC} \\
\midrule
\rowcolor{Gray}\multicolumn{3}{c}{\textit{\textbf{GPT-4o}}} \\
All-at-Once   & 53.44 & \textcolor{gray}{\best{59.77}} \\
Step-by-Step  & 30.16 & 49.43 \\
Binary Search & \textcolor{gray}{\best{58.73}} & 12.64 \\
CORRECT       & 54.50 & 54.02 \\
CHIEF         & 40.21 & 48.28 \\
ErrorProbe    & 56.08 & 55.17 \\
\midrule
\rowcolor{Gray}\multicolumn{3}{c}{\textit{\textbf{Qwen3.5-9B}}} \\
All-at-Once   & 56.61 & 48.28 \\
Step-by-Step  & 19.05 & 47.13 \\
Binary Search & 38.10 & 62.07 \\
CORRECT       & 59.79 & 47.13 \\
CHIEF         & 50.26 & 54.02 \\
ErrorProbe    & 53.44 & 65.52 \\
StepFinder    & 39.26 & 37.01 \\
OAT           & 40.53 & 53.10 \\
\rowcolor{gray!10}\soap{}       & \best{60.32} & \best{67.82} \\
\midrule
\rowcolor{Gray}\multicolumn{3}{c}{\textit{\textbf{DeepSeek-R1-Distill-Llama-8B}}} \\
All-at-Once   & 42.33 & 45.98 \\
Step-by-Step  & 44.97 & 37.93 \\
Binary Search & 28.57 & 48.28 \\
CORRECT       & 44.97 & 32.18 \\
CHIEF         & 37.04 & 52.87 \\
ErrorProbe    & 43.39 & 47.13 \\
StepFinder    & 48.99 & 41.15 \\
OAT           & 42.33 & 56.32 \\
\rowcolor{gray!10}\soap{}       & \best{61.90} & \best{65.52} \\
\bottomrule
\end{tabular}
\end{table}

\noindent \textbf{Variance across seeds.}
Table~\ref{tab:std} gives the standard deviation over the three seeds of every
step-level number in Table~\ref{tab:main}. On the small subsets one trajectory is
worth 1.6 points (\textsc{WW-AG}, 63 test trajectories) to 3.4 points (\textsc{WW-HC}, 29), so standard deviations of 2--5 points are the norm for every method, and differences
below that on a single subset should not be read; \textsc{CE}, with over a thousand test
trajectories, is stable to under one point. \soap{}'s margin over the strongest
prompt-based baseline on \textsc{WW-AG} (47.62 vs.\ 39.68 for ErrorProbe with Qwen3.5-9B) is
about three standard deviations, and its margins over the training-based methods
and over every judge on DeepSeek-8B are larger still.

\begin{table}[h]
\centering
\footnotesize
\setlength{\tabcolsep}{3.5pt}
\caption{Step-level accuracy (\%) in the without-ground-truth setting as mean $\pm$
standard deviation over the three seeds of the frozen triple, for every row of
Table~\ref{tab:main}. OAT and StepFinder are first averaged over
their five training seeds within each split.}
\label{tab:std}
\begin{tabular}{l ccccc}
\toprule
Method & \textsc{WW-AG} & \textsc{WW-HC} & \textsc{CE} & \textsc{TE-Cap} & \textsc{TE-Mag} \\
\midrule
\rowcolor{Gray}\multicolumn{6}{c}{\textit{\textbf{GPT-4o}}} \\
All-at-Once   & $\ms{14.81}{2.70}$ & $\ms{6.90}{0.00}$  & $\ms{28.39}{0.36}$ & $\ms{16.28}{3.29}$ & $\ms{5.80}{2.05}$ \\
Step-by-Step  & $\ms{20.63}{1.30}$ & $\ms{13.79}{5.63}$ & $\ms{44.47}{1.25}$ & $\ms{13.95}{6.85}$ & $\ms{13.04}{3.55}$ \\
Binary Search & $\ms{22.22}{3.43}$ & $\ms{4.60}{1.63}$  & $\ms{9.70}{0.82}$  & $\ms{9.30}{1.90}$  & $\ms{6.52}{0.00}$ \\
CORRECT       & $\ms{15.34}{0.75}$ & $\ms{4.60}{1.63}$  & $\ms{35.12}{1.93}$ & $\ms{10.85}{4.78}$ & $\ms{13.77}{2.71}$ \\
CHIEF         & $\ms{25.93}{3.96}$ & $\ms{14.94}{1.63}$ & $\ms{38.56}{0.86}$ & $\ms{13.95}{1.90}$ & $\ms{8.70}{3.07}$ \\
ErrorProbe    & \textcolor{gray}{{\boldmath$\ms{38.62}{1.98}$}} & \textcolor{gray}{{\boldmath$\ms{21.84}{3.25}$}} & \textcolor{gray}{{\boldmath$\ms{59.52}{1.49}$}} & \textcolor{gray}{{\boldmath$\ms{27.13}{2.19}$}} & \textcolor{gray}{{\boldmath$\ms{28.26}{3.07}$}} \\
\midrule
\rowcolor{Gray}\multicolumn{6}{c}{\textit{\textbf{Qwen3.5-9B}}} \\
All-at-Once   & $\ms{17.46}{2.59}$ & $\ms{11.49}{4.30}$ & $\ms{33.47}{0.53}$ & $\ms{4.65}{1.90}$  & $\ms{1.45}{2.05}$ \\
Step-by-Step  & $\ms{10.58}{1.50}$ & $\ms{16.09}{1.63}$ & $\ms{31.15}{2.46}$ & $\ms{14.73}{2.90}$ & $\ms{10.87}{4.70}$ \\
Binary Search & $\ms{2.65}{0.75}$  & $\ms{13.79}{5.63}$ & $\ms{59.92}{0.88}$ & $\ms{0.78}{1.10}$  & $\ms{5.07}{3.69}$ \\
CORRECT       & $\ms{29.10}{2.99}$ & $\ms{3.45}{0.00}$  & $\ms{36.35}{0.46}$ & $\ms{7.75}{1.10}$  & $\ms{2.90}{2.71}$ \\
CHIEF         & $\ms{28.57}{2.24}$ & $\ms{3.45}{0.00}$  & $\ms{29.39}{0.65}$ & $\ms{3.88}{1.10}$  & $\ms{2.17}{1.77}$ \\
ErrorProbe    & $\ms{39.68}{2.24}$ & $\ms{21.84}{5.86}$ & $\ms{55.03}{1.21}$ & $\ms{23.26}{3.80}$ & $\ms{20.29}{2.71}$ \\
StepFinder    & $\ms{15.87}{0.78}$ & $\ms{13.33}{3.39}$ & $\ms{30.03}{1.07}$ & $\ms{18.29}{3.90}$ & $\ms{6.67}{0.54}$ \\
OAT           & $\ms{16.72}{0.79}$ & $\ms{10.11}{6.33}$ & $\ms{56.50}{0.82}$ & $\ms{15.35}{1.52}$ & $\ms{18.84}{2.76}$ \\
\rowcolor{gray!10}\soap{}       & {\boldmath$\ms{47.62}{2.59}$} & {\boldmath$\ms{34.48}{2.82}$} & {\boldmath$\ms{61.78}{0.73}$} & {\boldmath$\ms{35.66}{1.10}$} & {\boldmath$\ms{23.19}{4.47}$} \\
\midrule
\rowcolor{Gray}\multicolumn{6}{c}{\textit{\textbf{DeepSeek-R1-Distill-Llama-8B}}} \\
All-at-Once   & $\ms{7.41}{3.26}$  & $\ms{5.75}{1.63}$  & $\ms{21.95}{0.85}$ & $\ms{4.65}{1.90}$  & $\ms{2.90}{1.02}$ \\
Step-by-Step  & $\ms{16.93}{1.98}$ & $\ms{9.20}{1.63}$  & $\ms{9.04}{1.74}$  & $\ms{11.63}{1.90}$ & $\ms{4.35}{1.77}$ \\
Binary Search & $\ms{7.94}{1.30}$  & $\ms{12.64}{3.25}$ & $\ms{34.15}{0.94}$ & $\ms{10.08}{5.80}$ & $\ms{13.77}{2.71}$ \\
CORRECT       & $\ms{20.11}{1.98}$ & $\ms{11.49}{3.25}$ & $\ms{32.82}{1.12}$ & $\ms{3.88}{1.10}$  & $\ms{0.72}{1.02}$ \\
CHIEF         & $\ms{17.46}{3.43}$ & $\ms{2.30}{1.63}$  & $\ms{8.40}{0.42}$  & $\ms{7.75}{1.10}$  & $\ms{9.42}{2.71}$ \\
ErrorProbe    & $\ms{24.34}{1.98}$ & $\ms{10.34}{2.82}$ & $\ms{35.97}{0.41}$ & $\ms{17.05}{2.90}$ & $\ms{8.70}{1.77}$ \\
StepFinder    & $\ms{18.94}{2.84}$ & $\ms{10.80}{3.10}$ & $\ms{36.91}{0.22}$ & $\ms{14.73}{0.79}$ & $\ms{4.20}{1.08}$ \\
OAT           & $\ms{12.70}{2.99}$ & $\ms{15.86}{8.29}$ & $\ms{52.50}{1.13}$ & $\ms{17.98}{0.22}$ & $\ms{13.04}{1.98}$ \\
\rowcolor{gray!10}\soap{}       & {\boldmath$\ms{45.50}{2.70}$} & {\boldmath$\ms{28.74}{4.30}$} & {\boldmath$\ms{64.68}{0.42}$} & {\boldmath$\ms{42.64}{4.39}$} & {\boldmath$\ms{30.43}{3.07}$} \\
\bottomrule
\end{tabular}
\end{table}

\subsection{Accuracy at $k$}
\label{app:ranked}

\soap{}, OAT, and StepFinder each produce a per-step score, so the top-$k$ steps form a ranked shortlist for triage; the prompt-based judges return a single verdict and are excluded. Table~\ref{tab:ranked-step} reports step-level accuracy at $k \in \{1, 3, 5\}$ and mean reciprocal rank (MRR) on the test splits of Table~\ref{tab:main}. For OAT and StepFinder we rank their stored scores; a step that OAT never scored counts as a miss at every $k$.

\soap{} holds its lead as the shortlist grows. On \textsc{WW-AG} with Qwen3.5-9B, it locates the decisive step in 74.60\% of trajectories at $k{=}3$, against 56.72 for OAT and 55.24 for StepFinder, and in 91.01\% at $k{=}5$. On \textsc{WW-HC}, whose trajectories average 52 turns, the $k{=}5$ gap is 12 points over OAT (50.57 vs.\ 38.39). The baselines close in on \textsc{Correct-Error}, where five candidates already cover most of a trajectory (Table~\ref{tab:datasets}). MRR confirms the ranking: 63.92 on \textsc{WW-AG} with Qwen3.5-9B, against 41.46 for OAT and 41.23 for StepFinder.

\begin{table}[h]
\centering
\footnotesize
\setlength{\tabcolsep}{3pt}
\caption{Step-level accuracy at $k$ (\%) and mean reciprocal rank (MRR, \%) in the
without-ground-truth setting for the three methods that rank every step. The best
result in each column within a block is highlighted in \textbf{bold}.}
\label{tab:ranked-step}
\resizebox{\ifdim\width>\textwidth \textwidth\else\width\fi}{!}{%
\begin{tabular}{l ccc ccc ccc ccc ccc c}
\toprule
 & \multicolumn{3}{c}{\textsc{WW-AG}} & \multicolumn{3}{c}{\textsc{WW-HC}} & \multicolumn{3}{c}{\textsc{CE}} & \multicolumn{3}{c}{\textsc{TE-Cap}} & \multicolumn{3}{c}{\textsc{TE-Mag}} & MRR \\
\cmidrule(lr){2-4}\cmidrule(lr){5-7}\cmidrule(lr){8-10}\cmidrule(lr){11-13}\cmidrule(lr){14-16}
Method & @1 & @3 & @5 & @1 & @3 & @5 & @1 & @3 & @5 & @1 & @3 & @5 & @1 & @3 & @5 & \textsc{WW-AG} \\
\midrule
\rowcolor{Gray}\multicolumn{17}{c}{\textit{\textbf{Qwen3.5-9B}}} \\
OAT        & 16.72 & 56.72 & 86.46 & 10.11 & 25.75 & 38.39 & 56.50 & 74.60 & 82.18 & 15.35 & 53.02 & 73.80 & 18.84 & 32.17 & 40.29 & 41.46 \\
StepFinder & 15.87 & 55.24 & 75.87 & 13.33 & 29.66 & 34.48 & 30.03 & 59.04 & 68.58 & 18.29 & 37.21 & 50.70 & 6.67 & 17.10 & 28.12 & 41.23 \\
\rowcolor{gray!10}\soap{}    & \best{47.62} & \best{74.60} & \best{91.01} & \best{34.48} & \best{42.53} & \best{50.57} & \best{61.78} & \best{79.61} & \best{85.74} & \best{35.66} & \best{63.57} & \best{75.97} & \best{23.19} & \best{32.61} & \best{42.75} & \best{63.92} \\
\midrule
\rowcolor{Gray}\multicolumn{17}{c}{\textit{\textbf{DeepSeek-R1-Distill-Llama-8B}}} \\
OAT        & 12.70 & 45.19 & 83.92 & 15.86 & 37.01 & 42.30 & 52.50 & 74.08 & 83.17 & 17.98 & 47.13 & 75.04 & 13.04 & 31.01 & 41.45 & 37.62 \\
StepFinder & 18.94 & 60.32 & 78.94 & 10.80 & 21.61 & 29.20 & 36.91 & 63.05 & 74.12 & 14.73 & 41.24 & 58.60 & 4.20 & 11.01 & 21.45 & 43.75 \\
\rowcolor{gray!10}\soap{}    & \best{45.50} & \best{76.19} & \best{93.12} & \best{28.74} & \best{42.53} & \best{59.77} & \best{64.68} & \best{80.30} & \best{87.41} & \best{42.64} & \best{62.02} & \best{78.29} & \best{30.43} & \best{39.13} & \best{44.20} & \best{63.71} \\
\bottomrule
\end{tabular}}
\end{table}

\subsection{Results with GPT-5 as the judge}
\label{app:prompting-gpt5}

Table~\ref{tab:gpt5} repeats the prompt-based rows of Tables~\ref{tab:main} and
\ref{tab:main-gt} with GPT-5 as the judge on \textsc{Who\&When} and \textsc{TraceElephant}; the \soap{} rows are those of the main tables. We leave out \textsc{Correct-Error}: its 2{,}226 trajectories and the multi-call protocols make it too expensive to run with GPT-5. GPT-5 lifts every judge, Binary Search most (41.80 on \textsc{WW-AG} without the answer) and ErrorProbe on \textsc{WW-AG} (49.21), so the judge's capacity rather than the prompting protocol sets these methods' accuracy. Even so, \soap{} on a 9B or 8B backbone stays ahead of every GPT-5 judge on \textsc{WW-HC}, \textsc{TE-Cap}, and (on DeepSeek-8B) \textsc{TE-Mag} without the answer; on \textsc{WW-AG}, ErrorProbe with GPT-5 overtakes it (49.21 vs.\ 47.62 without the answer, 53.44 vs.\ 43.39 with it). These results indicate that a stronger judge narrows the gap on the short algorithm-generated trajectories and in the answer-aided setting, while \soap{} remains the most accurate method on the long agentic trajectories with no judge at all.

\begin{table}[h]
\centering
\footnotesize
\setlength{\tabcolsep}{6pt}
\caption{Step-level accuracy (\%) of the prompt-based methods with GPT-5 as the
judge, without and with the ground-truth answer, on \textsc{Who\&When} and \textsc{TraceElephant}.
\textsc{Correct-Error} was not run with GPT-5. The \soap{} rows repeat
Tables~\ref{tab:main} and \ref{tab:main-gt}. The best result in each column within
a block is highlighted in \textbf{bold}.}
\label{tab:gpt5}
\begin{tabular}{l cccc}
\toprule
Method & \textsc{WW-AG} & \textsc{WW-HC} & \textsc{TE-Cap} & \textsc{TE-Mag} \\
\midrule
\rowcolor{Gray}\multicolumn{5}{c}{\textit{\textbf{Without GT}}} \\
All-at-Once   & 18.52 & 4.60  & 10.85 & 10.87 \\
Step-by-Step  & 24.34 & 18.39 & 17.83 & 6.52 \\
Binary Search & 41.80 & 21.84 & 24.03 & 20.29 \\
CORRECT       & 24.87 & 13.79 & 20.93 & 23.91 \\
CHIEF         & 27.51 & 25.29 & 22.48 & 16.67 \\
ErrorProbe    & \best{49.21} & 16.09 & 26.36 & 20.29 \\
\rowcolor{gray!10}\soap{}, Qwen3.5-9B  & 47.62 & \best{34.48} & 35.66 & 23.19 \\
\rowcolor{gray!10}\soap{}, DeepSeek-8B & 45.50 & 28.74 & \best{42.64} & \best{30.43} \\
\midrule
\rowcolor{Gray}\multicolumn{5}{c}{\textit{\textbf{With GT}}} \\
All-at-Once   & 17.99 & 9.20  & 10.85 & 13.77 \\
Step-by-Step  & 28.04 & 24.14 & 20.16 & 15.22 \\
Binary Search & 47.09 & 24.14 & 31.01 & 25.36 \\
CORRECT       & 32.80 & 21.84 & 27.13 & 20.29 \\
CHIEF         & 29.63 & 18.39 & 14.73 & 22.46 \\
ErrorProbe    & \best{53.44} & 17.24 & 29.46 & 19.57 \\
\rowcolor{gray!10}\soap{}, Qwen3.5-9B  & 43.39 & \best{34.48} & 36.43 & 21.01 \\
\rowcolor{gray!10}\soap{}, DeepSeek-8B & 44.97 & 29.89 & \best{42.64} & \best{36.96} \\
\bottomrule
\end{tabular}
\end{table}

\subsection{Larger backbones}
\label{app:proxies}

Table~\ref{tab:scale} reports the numbers behind Figure~\ref{fig:scale}(a--b) with seed standard deviations and baselines at every size. Within the Qwen3.5 family, \soap{} holds its level from 9B to 27B: 47.62 vs.\ 44.97 on \textsc{WW-AG} (a gap of about one standard deviation) and 34.48 on \textsc{WW-HC} at both sizes. The Qwen3-14B point is competitive on \textsc{WW-AG} (42.86) but 9 points lower on \textsc{WW-HC}; because it also crosses model generations, this gap reflects a family effect as much as a size effect. Neither baseline scales with model size. OAT stays within 14--20 on \textsc{WW-AG} and 10--17 on \textsc{WW-HC}, and StepFinder moves without a trend, with training-seed standard deviations of 1--6 points. The margin of Table~\ref{tab:main} holds at every size we tested.

\begin{table}[h]
\centering
\small
\setlength{\tabcolsep}{5pt}
\caption{Step-level accuracy (\%) by backbone size on \textsc{Who\&When}, mean $\pm$ standard
deviation (over the seed triple for \soap{}; over the five training seeds for OAT and
StepFinder). Qwen3-14B belongs to the previous model generation. The best result in
each column is highlighted in \textbf{bold}.}
\label{tab:scale}
\begin{tabular}{l l ccc}
\toprule
Subset & Method & Qwen3.5-9B & Qwen3-14B & Qwen3.5-27B \\
\midrule
\multirow{3}{*}{\textsc{WW-AG}}
 & OAT        & $\ms{16.72}{1.56}$ & $\ms{13.97}{1.48}$ & $\ms{20.42}{1.40}$ \\
 & StepFinder & $\ms{15.87}{2.99}$ & $\ms{22.65}{3.72}$ & $\ms{13.54}{2.08}$ \\
 & \cellcolor{gray!10}\soap{}    & \cellcolor{gray!10}{\boldmath$\ms{47.62}{2.59}$} & \cellcolor{gray!10}{\boldmath$\ms{42.86}{4.67}$} & \cellcolor{gray!10}{\boldmath$\ms{44.97}{2.70}$} \\
\midrule
\multirow{3}{*}{\textsc{WW-HC}}
 & OAT        & $\ms{10.11}{1.13}$ & $\ms{16.78}{2.00}$ & $\ms{12.41}{1.13}$ \\
 & StepFinder & $\ms{13.33}{4.80}$ & $\ms{14.02}{2.94}$ & $\ms{16.55}{6.24}$ \\
 & \cellcolor{gray!10}\soap{}    & \cellcolor{gray!10}{\boldmath$\ms{34.48}{2.82}$} & \cellcolor{gray!10}{\boldmath$\ms{25.29}{6.50}$} & \cellcolor{gray!10}{\boldmath$\ms{34.48}{2.82}$} \\
\bottomrule
\end{tabular}
\end{table}

\subsection{Cross-distribution transfer}
\label{app:transfer}

\begin{figure}[h]
\centering
\includegraphics[width=0.9\textwidth]{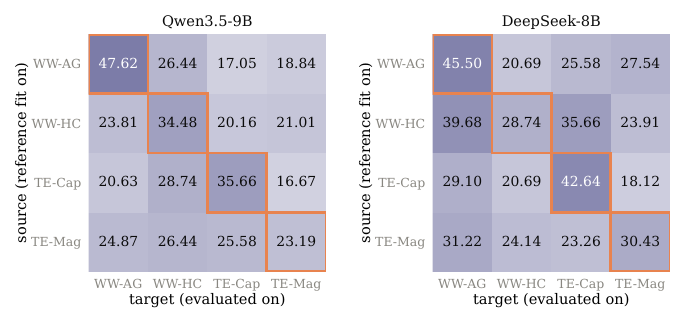}
\caption{Cross-distribution transfer, all grids. Step-level accuracy (\%) of \soap{}
with the reference fitted on the source subset (rows) and evaluated on the target subset (columns), for both backbones. The outlined
diagonal is the in-distribution number of Table~\ref{tab:main} in every grid.}
\label{fig:transfer-app}
\end{figure}

Figure~\ref{fig:transfer-app} gives the complete transfer grids for both backbones; the Qwen3.5-9B grid repeats Figure~\ref{fig:transfer}(c). For each source--target pair, the reference $R$ is fitted on the source's reference split, and the configuration is re-selected on the target's validation split, which for the cross setting pools the target's reference and validation files, since the target's reference split is not used for fitting; the reported number is the resulting test accuracy. The diagonal of each grid is the in-distribution number of Table~\ref{tab:main}. The diagonal wins every column on both backbones, so a reference fitted on the target's own distribution remains the best choice. The degradation off the diagonal, however, is graded, not catastrophic: DeepSeek-8B \textsc{WW-HC}$\to$\textsc{WW-AG} keeps 39.68 of the in-distribution 45.50, and \textsc{WW-HC}$\to$\textsc{TE-Cap} keeps 35.66 of 42.64.

\subsection{Synthetic references, full results}
\label{app:synth-results}

Table~\ref{tab:synth-app} extends Figure~\ref{fig:transfer}(d) to DeepSeek-8B (generation details in Appendix~\ref{app:synth}). On \textsc{WW-AG}, the existing benchmark reference remains strongest (45.50), while both synthetic corpora stay within 6--7 points (39.68 and 38.10). On \textsc{WW-HC} the Qwen3.5-9B-generated corpus slightly exceeds the existing benchmark reference (29.89 vs.\ 28.74). Neither generator consistently dominates: GPT-4o wins on \textsc{WW-AG} and Qwen3.5-9B wins on \textsc{WW-HC}, matching the pattern observed with the Qwen3.5-9B backbone in Figure~\ref{fig:transfer}(d). All synthetic rows remain well above the training-based baselines of Table~\ref{tab:main}. 

\begin{table}
\vspace{-0.3\baselineskip}
\centering
\footnotesize
\setlength{\tabcolsep}{4.5pt}
\renewcommand{\arraystretch}{1.04}
\caption{\small \textbf{Synthetic references with DeepSeek-8B.}
Step-level accuracy (\%) of \soap{}; only the reference corpus changes.}
\vspace{-0.3cm}
\label{tab:synth-app}
\begin{tabular}{l cc}
\toprule
Reference & \textsc{WW-AG} & \textsc{WW-HC} \\
\midrule
Existing benchmarks          & \best{45.50} & 28.74 \\
Synthetic, Qwen3.5-9B & 39.68 & \best{29.89} \\
Synthetic, GPT-4o      & 38.10 & 25.29 \\
\bottomrule
\end{tabular}
\end{table}

\subsection{Qualitative examples}
\label{app:qualitative}

Figures~\ref{fig:qual-text-1}--\ref{fig:qual-text-4} show four textual trajectories
in the test set, all scored with Qwen3.5-9B under the configuration selected
for Table~\ref{tab:main}. We denote step $i$ as $S_i$. Each step shows its agent, $[\Sbase \rightarrow \Stilde]$ (base and rescored score, $\times 10^{-3}$), and its content condensed by us; tags mark the annotated decisive step and each score's argmax. The third example (Figure~\ref{fig:qual-text-3}) is the trajectory whose score plot appears in Figure~\ref{fig:qualitative}. Agent names and step indices are as released. The \textsc{WW-HC} trajectory (Figure~\ref{fig:qual-text-2}) uses Magentic-One, whose orchestrator alternates between private ledger entries and dispatches: \textsf{(thought)} marks a ledger entry where the orchestrator reasons about progress, and $\rightarrow$ marks an instruction sent to an agent.

The four trajectories fail in different ways, but the base score misreads them in the same way: it flags a late symptom rather than the step that caused it. \soap{} rescores each step by accounting for the attention its successors pay to it, correcting this downstream contamination. For each example we give the task, the annotated decisive step and what went wrong, where the base score peaks, and where \soap{} lands.
\begin{itemize}
    \item \textbf{\textsc{WW-AG}} (Figure~\ref{fig:qual-text-1}). \emph{Task:} identify the command clicked in the last video of a 2018 VSCode blog post on replit.com. \emph{Decisive step, $S_1$:} the video-analysis expert submits code that loops over a \texttt{None} return from \texttt{perform\_web\_search}; the loop crashes immediately. \emph{Base score:} peaks at $S_2$, the \texttt{TypeError} traceback. \emph{\soap{}:} lifts $S_1$, the buggy call that produced the crash.

    \item \textbf{\textsc{WW-HC}} (Figure~\ref{fig:qual-text-2}). \emph{Task:} price mailing a DVD to Colombia with three carriers. \emph{Decisive step, $S_8$:} the WebSurfer ignores the instruction to look up DHL, clicks a FedEx rate tool, and returns an empty form with no rate. The orchestrator records FedEx as done, and every later turn builds on that belief. \emph{Base score:} peaks at $S_{73}$, the final search dump after 65 turns of loops and replans, a symptom at the far end of the contamination. \emph{\soap{}:} lifts $S_8$ above it, because the turns that stalled on the FedEx form attend to $S_8$.

    \item \textbf{\textsc{TE-Cap}} (Figure~\ref{fig:qual-text-3}; score plot in Figure~\ref{fig:qualitative}). \emph{Task:} identify the missing character in an Unlambda program that should output ``For penguins''. \emph{Decisive step, $S_1$:} the extraction expert misanalyzes the code's combinator structure, concludes a dot is needed before \texttt{s}, and answers ``.'' (the correct answer is ``backtick''). \emph{Base score:} peaks at $S_5$, the orchestrator's final restatement of the wrong answer. \emph{\soap{}:} moves the top-1 to $S_1$, because $S_3$--$S_5$ attend to the analysis at $S_1$.

    \item \textbf{\textsc{TE-Mag}} (Figure~\ref{fig:qual-text-4}). \emph{Task:} solve a game-show puzzle by code. \emph{Decisive step, $S_6$:} the orchestrator fixes a conservative strategy, the same guess for every box. A \texttt{NameError} follows at $S_8$, and at $S_9$ the orchestrator re-issues the script with the bug fixed and the strategy unchanged. \emph{Base score:} peaks at $S_9$, the bug fix. \emph{\soap{}:} lifts $S_6$, on which both the crash and the fix depend.
\end{itemize}

\newcommand{\qsc}[2]{{\color{black!60}\scriptsize[#1\,$\rightarrow$\,#2]}}
\newcommand{\qtag}[2]{{\setlength{\fboxsep}{1pt}\colorbox{#1}{\scriptsize\textbf{#2}}}}
\newcommand{\qgold}{\qtag{green!25}{decisive}}
\newcommand{\qbase}{\qtag{red!20}{base top-1}}
\newcommand{\qsoap}{\qtag{blue!18}{\soap{} top-1}}
\newcommand{\qstep}[4]{\par\noindent\hangindent=1.3em\hangafter=1$\mathbf{S_{\text{#1}}}$\ {\color{black!65}\textsf{#2}}\ \qsc{#3}{#4}\ }
\newcommand{\qsub}[4]{\par\noindent\hangindent=1.3em\hangafter=1$\mathbf{S_{\text{#1}}}$\ {\color{black!65}\textsf{#2}}\ \qsc{#3}{#4}\ }
\newcommand{\qhead}[1]{\par\noindent\textbf{#1}\ }
\newtcolorbox{qbox}[1]{colback=black!2,colframe=black!55,boxrule=0.5pt,arc=0pt,
  left=4pt,right=4pt,top=3pt,bottom=3pt,
  title={#1},fonttitle=\bfseries\footnotesize,colbacktitle=black!8,coltitle=black,
  titlerule=0pt,toptitle=1.5pt,bottomtitle=1.5pt,before upper=\sloppy}

\begin{figure}[ht]
\centering
\begin{minipage}{\textwidth}
\begin{qbox}{\textsc{WW-AG} \,$\cdot$\, CaptainAgent}
\fontsize{7.4}{8.6}\selectfont
\qhead{Task.} In the 2018 VSCode blog post on replit.com, what was the command they
clicked on in the last video to remove extra lines?
\qhead{Gold answer.} Format Document.
\qhead{Annotation.} $S_1$, VideoContentAnalysis\_Expert: ``The code is incorrect for the task.''
\qhead{Prediction.} Base score $S_2$; \soap{} $S_1$.
\medskip
\qstep{0}{VSCode\_Expert}{12.32}{19.94} Restates the manager's brief.
\qstep{1}{VideoContentAnalysis\_Expert}{17.44}{30.42}\qgold\ \qsoap\ Submits code that
loops over \texttt{perform\_web\_search}'s \texttt{None} return. The loop fails.
\qstep{2}{Computer\_terminal}{21.64}{21.64}\qbase\ \texttt{TypeError}: \texttt{'NoneType'} not iterable.
\qstep{3}{VideoContentAnalysis\_Expert}{15.76}{28.63} Asks the group to open the link.
\qstep{4}{VideoContentAnalysis\_Expert}{21.45}{21.45} Repeats the instructions.
\qstep{5}{VideoContentAnalysis\_Expert}{4.87}{4.87} TERMINATE. No answer.
\end{qbox}
\end{minipage}
\caption{\textsc{WW-AG} CaptainAgent trajectory, scored with Qwen3.5-9B.}
\label{fig:qual-text-1}
\end{figure}

\begin{figure}[p]
\centering
\begin{minipage}{\textwidth}
\begin{qbox}{\textsc{WW-HC} \,$\cdot$\, Magentic-One}
\fontsize{6.4}{7.2}\selectfont
\qhead{Task.} What is the cheapest option to mail a DVD to Colombia from Hartford,
Connecticut using FedEx, DHL, or USPS? (Answer as JSON with keys \texttt{sender} and
\texttt{price (usd)}.) \textbf{Gold answer.} USPS, 41.75 USD.
\qhead{Annotation.} $S_8$, WebSurfer: ``did not return the useful information needed to
answer the question.'' \textbf{Prediction.} Base score $S_{73}$; \soap{} $S_8$.
\qhead{Reading.} $S_0$ is the unscored user question. \textsf{(thought)} marks the
orchestrator's ledger to itself, \textsf{$\rightarrow$} its instruction to an agent.
\smallskip

\qstep{1}{Orchestrator (thought)}{1.74}{1.88}
Initial plan: fact sheet (a DVD, Hartford to Colombia, three carriers); rates to look
up per carrier; team Assistant, ComputerTerminal, FileSurfer, WebSurfer.

\qstep{2}{Orchestrator (thought)}{1.55}{1.69}
Ledger: nothing gathered; start with FedEx.

\qstep{3}{Orchestrator $\rightarrow$ WebSurfer}{1.49}{1.49}
Look up FedEx rates.

\qstep{4}{WebSurfer}{1.71}{1.90}
Types the query into Bing; returns a screenshot and page metadata.

\qstep{5}{Orchestrator (thought)}{1.82}{2.00}
Ledger: no rates yet; progress.

\qstep{6}{Orchestrator $\rightarrow$ WebSurfer}{1.57}{1.57}
Look up DHL rates.

\qstep{7}{Orchestrator (thought)}{0.92}{0.92}
Next speaker: WebSurfer.

\qstep{8}{WebSurfer}{2.59}{2.74}\qgold\ \qsoap\
Ignores the DHL instruction; clicks ``Calculate Shipping Rates -- FedEx''. Returns the
FedEx rate-tool landing page: screenshot, schema.org metadata, an empty form, no rate.

\qstep{9}{Orchestrator (thought)}{1.82}{1.95}
Ledger: ``we have gathered information for FedEx''; DHL and USPS pending. No FedEx rate
was ever read; every later turn builds on this belief.

\qstep{10}{Orchestrator $\rightarrow$ WebSurfer}{1.56}{1.56}
Look up DHL rates.

\qstep{11}{Orchestrator (thought)}{0.93}{0.93}
Next speaker: WebSurfer.

\qstep{12}{Orchestrator (thought)}{1.51}{1.68}
Ledger: in a loop, errors on FedEx and DHL, no progress; hand to Assistant.

\qstep{13}{Orchestrator $\rightarrow$ Assistant}{1.51}{1.51}
Get USPS rates, then fix FedEx and DHL.

\qstep{14}{Orchestrator (thought)}{1.59}{1.71}
Next speaker: Assistant.

\qstep{15}{Assistant}{1.45}{1.61}
Writes a two-step plan (``I'll visit the USPS website'') but has no browser; returns nothing.

\qstep{16}{Orchestrator (thought)}{1.69}{1.82}
Ledger: rates still missing; try USPS next.

\qstep{17}{Orchestrator $\rightarrow$ WebSurfer}{1.56}{1.56}
Look up USPS rates.

\qstep{18}{Orchestrator (thought)}{0.91}{1.02}
Next speaker: WebSurfer.

\qstep{19}{WebSurfer}{1.44}{1.61}
Still on the FedEx tool: types ``Hartford, Connecticut'' into \textit{From}.

\qstep{20}{Orchestrator (thought)}{1.63}{1.78}
Ledger: FedEx partially explored, DHL and USPS still pending.

\qstep{21}{Orchestrator $\rightarrow$ WebSurfer}{1.47}{1.47}
Continue with USPS.

\qstep{22}{Orchestrator (thought)}{0.94}{0.94}
Next speaker: WebSurfer.

\qstep{23}{WebSurfer}{1.62}{1.76}
Clicks the ``Hartford, Connecticut, USA'' suggestion.

\qstep{24}{Orchestrator (thought)}{1.57}{1.71}
Ledger: FedEx form partially filled; USPS and DHL incomplete.

\qstep{25}{Orchestrator $\rightarrow$ WebSurfer}{1.46}{1.46}
USPS rates, then revisit DHL.

\qstep{26}{Orchestrator (thought)}{0.95}{0.95}
Next speaker: WebSurfer.

\qstep{27}{WebSurfer}{1.43}{1.60}
Types ``Colombia'' into the FedEx \textit{To} field.

\qstep{28}{Orchestrator (thought)}{1.78}{1.93}
Ledger: USPS rates still needed; FedEx form in progress.

\qstep{29}{Orchestrator $\rightarrow$ WebSurfer}{1.51}{1.51}
Continue with USPS.

\qstep{30}{Orchestrator (thought)}{0.96}{1.07}
Next speaker: WebSurfer.

\qstep{31}{WebSurfer}{1.65}{1.82}
Clicks ``Colombia''.

\qstep{32}{Orchestrator (thought)}{1.79}{1.93}
Ledger: USPS and DHL rates still missing.

\qstep{33}{Orchestrator $\rightarrow$ WebSurfer}{1.58}{1.58}
Look up USPS rates.

\qstep{34}{Orchestrator (thought)}{0.96}{0.96}
Next speaker: WebSurfer.

\qstep{35}{WebSurfer}{1.76}{1.76}
Clicks the browser back button; the FedEx form reads From Hartford 06103, destination
Colombia; no rate.

\qstep{36}{Orchestrator (thought)}{1.58}{1.69}
Ledger: cheapest option not yet determined.

\qstep{37}{Orchestrator (thought)}{1.67}{1.79}
Stalled; replanning.

\qstep{38}{Orchestrator (thought)}{1.55}{1.72}
New plan: use the USPS International Shipping Calculator.

\qstep{39}{Orchestrator (thought)}{1.80}{1.96}
Ledger: request not yet addressed; proceed with new plan.

\qstep{40}{Orchestrator $\rightarrow$ WebSurfer}{1.79}{1.79}
Open the USPS calculator, enter the DVD details.

\qstep{41}{Orchestrator (thought)}{1.06}{1.16}
Next speaker: WebSurfer.

\qstep{42}{WebSurfer}{1.87}{2.02}
Bing search ``USPS International Shipping Calculator''; results-page metadata.

\qstep{43}{Orchestrator (thought)}{1.72}{1.86}
Ledger: on the right track; click the calculator link.

\qstep{44}{Orchestrator $\rightarrow$ WebSurfer}{1.50}{1.50}
Click the calculator link, enter the details.

\qstep{45}{Orchestrator (thought)}{1.05}{1.05}
Next speaker: WebSurfer.

\qstep{46}{WebSurfer}{1.52}{1.71}
Clicks ``Calculate a Price'': the USPS Retail Postage Price Calculator opens, with the
destination defaulting to United States.

\qstep{47}{Orchestrator (thought)}{2.00}{2.18}
Ledger: loop, but on the right page; instruction: ZIP 06101, Colombia,
$7.5\times5.3\times0.6$ in, 0.2 lb.

\qstep{48}{Orchestrator $\rightarrow$ WebSurfer}{1.83}{1.83}
Enter ZIP 06101, Colombia, DVD dimensions and weight.

\qstep{49}{Orchestrator (thought)}{1.01}{1.01}
Next speaker: WebSurfer.

\qstep{50}{WebSurfer}{1.53}{1.72}
Clicks ``International''.

\qstep{51}{Orchestrator (thought)}{1.80}{1.97}
Ledger: USPS rates not yet obtained.

\qstep{52}{Orchestrator $\rightarrow$ WebSurfer}{1.76}{1.76}
Same instruction.

\qstep{53}{Orchestrator (thought)}{1.09}{1.09}
Next speaker: WebSurfer.

\qstep{54}{WebSurfer}{2.04}{2.04}
Clicks ``International'' again.

\qstep{55}{Orchestrator (thought)}{1.81}{1.99}
Ledger: USPS rates not yet retrieved.

\qstep{56}{Orchestrator $\rightarrow$ WebSurfer}{1.79}{1.79}
Same instruction.

\qstep{57}{Orchestrator (thought)}{1.11}{1.11}
Next speaker: WebSurfer.

\qstep{58}{WebSurfer}{2.24}{2.46}
Types 06101 into the from-ZIP field; destination still United States.

\qstep{59}{Orchestrator (thought)}{1.98}{2.16}
Ledger: USPS rates still pending.

\qstep{60}{Orchestrator $\rightarrow$ WebSurfer}{1.88}{1.88}
Select Colombia, enter dimensions and weight.

\qstep{61}{Orchestrator (thought)}{1.14}{1.14}
Next speaker: WebSurfer.

\qstep{62}{WebSurfer}{1.72}{1.89}
Clicks the destination-country field.

\qstep{63}{Orchestrator (thought)}{1.79}{1.96}
Ledger: USPS rates not yet obtained.

\qstep{64}{Orchestrator $\rightarrow$ WebSurfer}{1.71}{1.71}
Select Colombia, enter dimensions and weight.

\qstep{65}{Orchestrator (thought)}{1.08}{1.08}
Next speaker: WebSurfer.

\qstep{66}{WebSurfer}{1.70}{1.70}
Clicks the field again; nothing selected.

\qstep{67}{Orchestrator (thought)}{1.83}{1.83}
Ledger: loop, no progress.

\qstep{68}{Orchestrator (thought)}{1.23}{1.23}
Stalled; replanning.

\qstep{69}{Orchestrator (thought)}{1.49}{1.67}
New plan: retry the USPS calculator from scratch.

\qstep{70}{Orchestrator (thought)}{1.86}{2.03}
Ledger: full answer not yet provided; proceed.

\qstep{71}{Orchestrator $\rightarrow$ WebSurfer}{1.70}{1.70}
Open the USPS calculator, fill every field.

\qstep{72}{Orchestrator (thought)}{0.97}{0.97}
Next speaker: WebSurfer.

\qstep{73}{WebSurfer}{2.65}{2.65}\qbase\
Types ``USPS Retail Postage Price Calculator'' into Bing once more; returns a
13k-character dump of result metadata. The run ends with no rate from any carrier.
\end{qbox}
\end{minipage}
\caption{\textsc{WW-HC} Magentic-One trajectory, scored with Qwen3.5-9B.}
\label{fig:qual-text-2}
\end{figure}

\begin{figure}[ht]
\centering
\begin{minipage}{\textwidth}
\begin{qbox}{\textsc{TE-Cap} \,$\cdot$\, CaptainAgent}
\fontsize{7.4}{8.6}\selectfont
\qhead{Task.} In Unlambda, what exact character or text needs to be added to correct
the following code to output ``For penguins''? If what is needed is a character, answer
with the name of the character.
Code: \texttt{`r```````````.F.o.r.\ .p.e.n.g.u.i.n.si}
\qhead{Gold answer.} backtick.
\qhead{Annotation.} $S_1$, InformationExtraction\_Expert: ``a missing backtick disrupted
function application, preventing the code from correctly constructing and outputting
`For penguins'.\!''
\qhead{Prediction.} Base score $S_5$; \soap{} $S_1$.
\medskip

\qstep{0}{Orchestrator}{0.75}{0.84}
Assembles the team: InformationExtraction\_Expert (web search and result parsing),
CSVHandling\_Expert (Python over local files), Verification\_Expert (re-runs checks
against cited evidence).

\qstep{1}{InformationExtraction\_Expert}{0.87}{0.95}\qgold\ \qsoap\
Analyzes the Unlambda code. Identifies the dot operators and backtick-based function
application, but miscounts the applications: concludes a dot is needed before
\texttt{s} to include it in the output string. Answers ``.'' (dot).

\qstep{2}{Orchestrator}{0.69}{0.78}
Next speaker: Verification\_Expert.

\qstep{3}{Verification\_Expert}{0.82}{0.90}
Agrees with the dot answer. Repeats the reasoning without independent verification of
the combinator structure.

\qstep{4}{Orchestrator}{0.85}{0.93}
Summarizes: the task required identifying the missing character, determined to be a
dot before \texttt{s}.

\qstep{5}{Orchestrator}{0.88}{0.88}\qbase\
Final ANSWER ``.'' with the same justification. TERMINATE. The correct answer is
``backtick''.
\medskip

\par\noindent The final answer ($S_5$) and the summary ($S_4$) are restatements; the correction lifts
the flawed analysis that produced them.
\end{qbox}
\end{minipage}
\caption{\textsc{TE-Cap} CaptainAgent trajectory (traj-59), scored with Qwen3.5-9B. This is the trajectory whose score plot appears in Figure~\ref{fig:qualitative}.}
\label{fig:qual-text-3}
\end{figure}

\begin{figure}[ht]
\centering
\begin{minipage}{\textwidth}
\begin{qbox}{\textsc{TE-Mag} \,$\cdot$\, Magentic-One}
\fontsize{7.4}{8.6}\selectfont
\qhead{Task.} A game-show host hides 30 coins (\$1,000 each) in three boxes; one box
holds at least 2 coins and one holds 6 more than another. Bob guesses a count for
each box and wins his guess whenever it does not exceed the box, otherwise nothing.
With optimal play, what is the minimum amount Bob can win?
\qhead{Gold answer.} 16,000.
\qhead{Annotation.} $S_6$, Orchestrator: ``assuming Bob should make the same conservative
guess for each box, which ignored optimizing guesses based on varying coin
distributions.''
\qhead{Prediction.} Base score $S_9$; \soap{} $S_6$ (on all three seeds of the triple).
\medskip

\qstep{0}{Orchestrator}{0.72}{1.37}
Fact sheet: the givens (30 coins, three boxes, the two placement rules, the payoff
rule); nothing to look up.

\qstep{1}{Orchestrator}{0.61}{1.22}
Plan: enumerate the valid distributions, derive the guess that maximizes the worst
case, implement it with Coder, report.

\qstep{2}{Coder}{0.55}{1.09}
Ledger: proceed to the distributions; not in a loop; progress is being made.

\qstep{3}{Orchestrator}{0.75}{1.44}
Writes \texttt{find\_valid\_distributions()}: loops over $(a,b,c)$ summing to 30
with $a\ge2$ and some pair differing by 6; prints the list.

\qstep{4}{Coder}{0.50}{1.05}
Ledger: run the script.

\qstep{5}{Coder}{0.52}{1.10}
Ledger: distributions obtained; next, the strategy code.

\qstep{6}{Orchestrator}{0.70}{1.44}\qgold\ \qsoap\
Writes \texttt{calculate\_min\_winnings()}: for every distribution sets all three
guesses to \texttt{min(a,b,c)} ``as a conservative strategy'', sums them, and takes
the minimum over distributions. The guesses ignore that Bob chooses per box, so the
answer collapses to three times the smallest box.

\qstep{7}{Coder}{0.48}{1.04}
Ledger: run the script.

\qstep{8}{Coder}{0.56}{1.15}
Ledger: execution failed, \texttt{NameError}: \texttt{valid\_distributions} is
undefined; carry the earlier result over; progress: false.

\qstep{9}{Orchestrator}{0.89}{1.41}\qbase\
Explains the \texttt{NameError} and re-issues both functions as one combined script;
the strategy is unchanged.

\qstep{10}{Coder}{0.52}{1.08}
Ledger: execute the combined script.

\qstep{11}{Coder}{0.56}{1.27}
Ledger: ``Bob can ensure winning 3 coins''; provide the answer.

\qstep{12}{Coder}{0.71}{0.71}
ANSWER \$3,000: the conservative guess always wins at least 3 coins. TERMINATE.
\medskip

\par\noindent The crash report ($S_8$) and the fix ($S_9$) are the symptoms of the strategy fixed in $S_6$.
\end{qbox}
\end{minipage}
\caption{\textsc{TE-Mag} Magentic-One trajectory, scored with Qwen3.5-9B.}
\label{fig:qual-text-4}
\end{figure}

\end{document}